\documentclass[a4paper,11pt]{article}
\usepackage{geometry}
\usepackage{amsmath}
\usepackage{amssymb}
\usepackage{caption}
\usepackage{graphicx}
\usepackage{latexsym}
\usepackage{times} 
\usepackage[english]{babel}
\usepackage{fancyhdr}
\usepackage{xcolor}
\usepackage{subcaption}
\usepackage{verbatim}

\fancyfootoffset{1cm}
\fancypagestyle{plain}{%
\fancyhead{}
\fancyfoot{}
\rfoot{\thepage}
\lfoot{{\small EVS30 International Battery, Hybrid and Fuel Cell Electric Vehicle Symposium}}

}

\renewenvironment{abstract}
 {
  {\bfseries \large{\abstractname}}
  \par
  \vspace{10pt}
  \normalsize
 }

\title{\Large \emph{EVS30 Symposium}\\ \emph{Stuttgart, Germany, October 9 - 11, 2017}\\ \hspace{10pt}\\ \LARGE\bf
Autonomous Electric Race Car Design}

\author{{\large            Niklas Funk$^1$, Nikhilesh Alatur$^1$, Robin Deuber$^1$, Frederick Gonon$^1$, Nico Messikommer$^1$}\\
{\large Julian Nubert$^1$, Moritz Patriarca$^1$, Simon Schaefer$^1$, Dominic Scotoni$^1$, Nicholas B\"unger$^1$}\\
{\large Renaud Dub\'e$^1$, Raghav Khanna$^1$, Mark Pfeiffer$^1$, Erik Wilhelm$^2$, Roland Siegwart$^1$}\\
        {\small $^1$\em ETH-Zurich, R\"amistrasse 101, 8092 Zurich}\\
        {\small $^2$\em Kyburz Switzerland AG, Shedweg 2-8, 8427 Freienstein-Teufen}}

\date{}
\begin{document}
\sloppy

\setlength{\parindent}{0mm}
\baselineskip 10pt

\twocolumn[
  \begin{@twocolumnfalse}
\maketitle

\rule{\textwidth}{1pt}
\begin{abstract}
Autonomous driving and electric vehicles are nowadays very active research and development areas. 
In this paper we present the conversion of a standard Kyburz eRod into an autonomous vehicle that can be operated in challenging environments such as Swiss mountain passes.
The overall hardware and software architectures are described in detail with a special emphasis on the sensor requirements for autonomous vehicles operating in partially structured environments.
Furthermore, the design process itself and the finalized system architecture are presented. 
The work shows state of the art results in localization and controls for self-driving high-performance electric vehicles. 
Test results of the overall system are presented, which show the importance of generalizable state estimation algorithms to handle a plethora of conditions.


\end{abstract}

\rule{\textwidth}{1pt}
\vspace{10pt}
  \end{@twocolumnfalse}
]

\section{Introduction}
Due to the vast advances in computing and sensing systems, work in autonomous driving has grown significantly within the last decade. 
While the idea of autonomous vehicles has been around for many years the DARPA grand challenge \cite{thrun2006stanley} was one of the first major forays into self-driving vehicles. 



The task of autonomous driving can be split into two main parts: (i) The perception, localization and state estimation and (ii) the motion planning and control task.
Regarding localization it is important that the vehicle can precisely estimate its current state w.r.t. the environment. 
The motion planning and control part assures that the vehicle behaves as expected and executes the planned actions reliably.

In this work we present an autonomous driving application based on a visual teach-and-repeat method as presented in \cite{furgale}.
During a demonstration (teaching) run, a human vehicle operator provides a demonstration of how he wants the vehicle to drive in a given environment. 
During that run, a visual feature map of the environment is built using a Simultaneous Localization and Mapping (SLAM) method and the vehicles' position in this map is recorded.
An alternative to the visual SLAM would be laser-based approaches as shown in \cite{krusi2016driving} or \cite{dube2016segmatch}. 
However, since camera-based systems have a significantly lower cost than laser-based ones, in this work we will focus on the approach \cite{vi-sensor} based on visual and inertial measurements.
In the repeat phase, the car localizes in the pre-recorded feature map and follows the teach path as close as possible using a reference tracking controller.

Although we can assume that the taught path is collision free we still have to ensure that the car operates safely at any time. 
Therefore, in addition to the stereo camera used for localization we use a laser-based obstacle detection system that allows to safely detect objects which are blocking the vehicles reference path.
The approach presented in this work only requires local sensing and has no need for continuous lane markings or a GNSS signal. 

The main contributions of this work are: 
\begin{itemize}
\item The conversion of a stock electric vehicle into an autonomous driving platform 
\item A visual teach-and-repeat pipeline based on state-of-the art localization and control techniques
\item Real-world tests with the autonomous platform in the challenging environment of a Swiss mountain pass.
\end{itemize}

The paper is structured as follows: In Section \ref{sec:system-overview} we give an overview of our platform. In Section \ref{sec:system-architecture} we present the architecture of the system and present the submodules. Section \ref{sec:safety} provides information about our safety measures, Section \ref{sec:results} presents the results before we conclude in Section \ref{sec:conclusion}.  

\section{ARC Overview}\label{sec:system-overview}


\begin{figure}[htbp!]
\centering
    \includegraphics[width=0.45\textwidth]{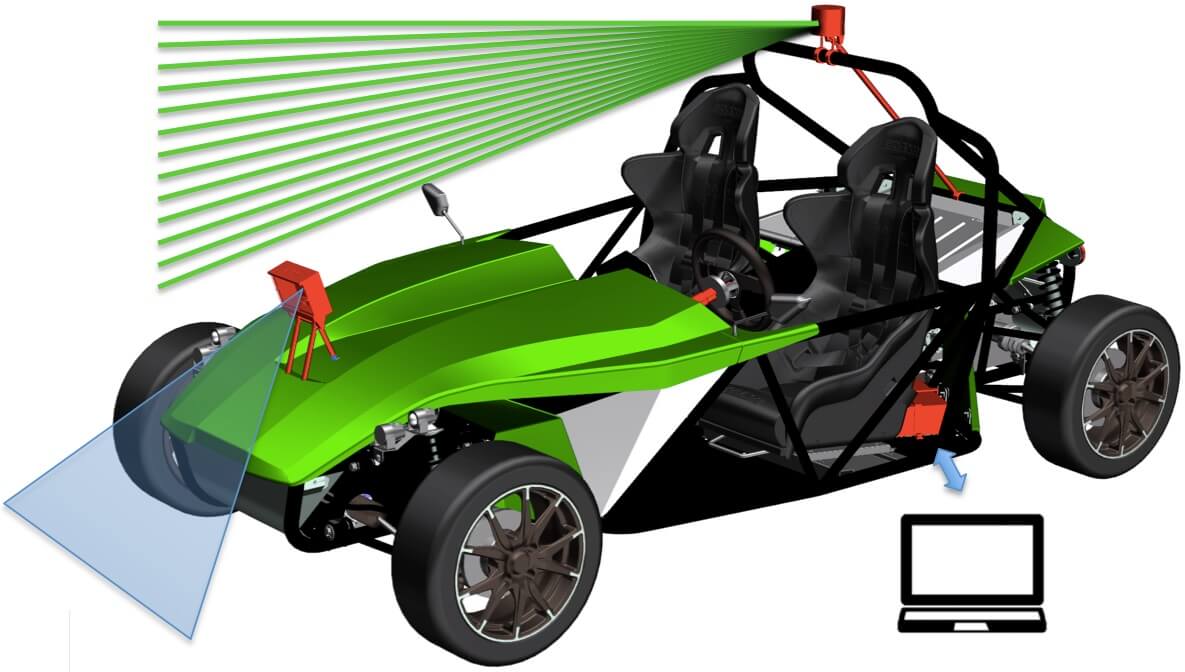}
    \caption{An illustration of our Autonomous eRod. The sensors, actuators and computer providing autonomous driving functionalities are shown in red.}
    \label{erod}
\end{figure}

The modified Kyburz eRod with a series of additional sensors is shown in Figure \ref{erod}. The eRod is an all-electric vehicle produced by Kyburz Switzerland and sold in European Markets and was provided to the ETH-Zurich final year student team for modification into a self-driving vehicle. The eRod has a nominal power of 15kW, and a peak power of 45kW, which, combined with its low weight of only 500kg, makes it a very nimble vehicle designed for fun and racing applications. 

In order to control the vehicle via digital commands, new actuators had to be deployed. Namely, a new steering unit provided by TKP which can be controlled via CAN was incorporated. Secondly, a new braking system was designed, consisting of an electric motor which pulls the braking pedal via metal cables and which can be controlled using an analog voltage input, and another voltage input which physically simulates the existing throttle potentiometer.


The choice of sensors is shown in Figure \ref{erod}. Our localization algorithms rely on the input from the Visual Inertial sensor (VI sensor) which is mounted on the bonnet. This sensor module contains two cameras together with an intertial measurement unit (IMU). The sensor is mounted in a way to allow adjustment of the sensor's angle around the pitch axis which is crucial since the view of the cameras has to be well-chosen. On the one hand it should not capture the sky since it is altering very quickly and therefore does not provide good features for re-localization. On the other hand, it should neither see the bonnet nor the vehicle's shadow which would confuse the vision based odometry estimate. An inclination angle of 11 degrees was found to optimally satisfy both of these criteria. 

The detection of obstacles relies on a Velodyne VLP-16 LiDAR sensor which is mounted on top of the roll bar. We selected a LiDAR sensor against a radar sensor as the latter can pose difficulties when detecting special surfaces and has a limited angular resolution\cite{bazer}. The VLP-16 was therefore chosen in order to allow an optimal obstacle detection and for enabling the investigation of future research avenues such as dynamic obstacle avoidance and LiDAR-based localization\cite{segmatch}.

The on-board sensor system also has two rotary wheel encoders which are read out using a FPGA on a  National Instruments compact RIO Vehicle Control Unit and a Steering Angle encoder which is read out using a dSpace Micro AutoBox II. The use of two seperate VCUs is not desirable from a complexity perspective, but saved lots of time due to the TKP steering unit having manufacturer-supplied SimuLink blocks and team familiarity with LabView.

\begin{figure}[htbp!]
\centering
    \includegraphics[width=0.45\textwidth]{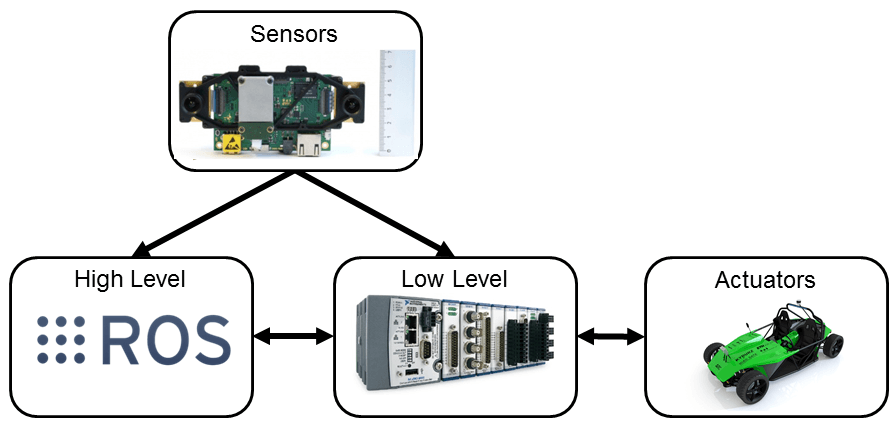}
    \caption{General system overview}
    \label{system}
\end{figure}

All the sensor information is fused on our Linux-based computer running ROS which completes the system showed in \ref{system}.

\begin{figure}[htbp!]
	\centering
	\begin{subfigure}{0.45\textwidth}
		\includegraphics[width=\textwidth]{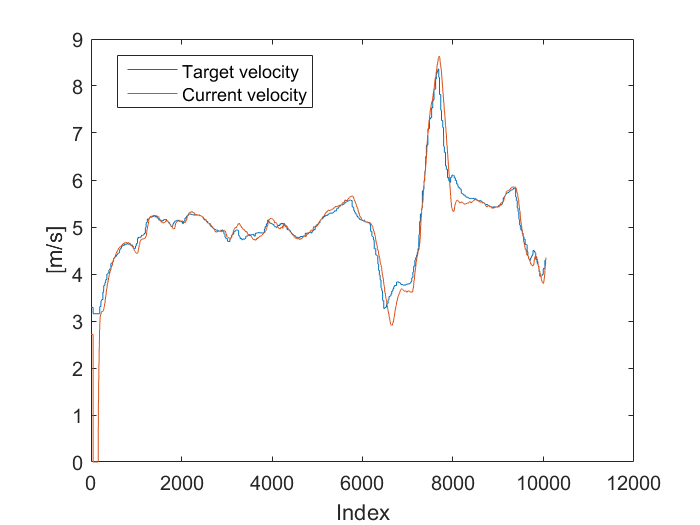}
	\end{subfigure}
	\begin{subfigure}{0.45\textwidth}
		\includegraphics[width=\textwidth]{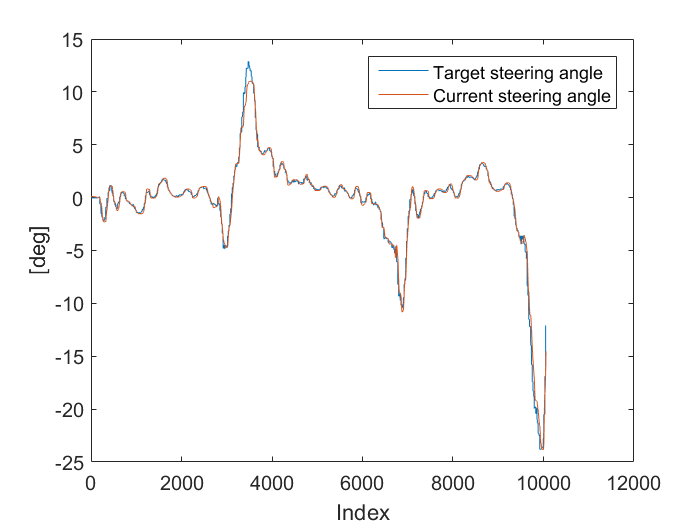}
	\end{subfigure}
	\caption{Low-level controller performance. Top: velocity controller. Bottom: steering controller.}
	\label{fig:low-level}
\end {figure}

The computer is the core of the autonomous system, taking care of the state estimation, obstacle detection, and finally the computation of the control commands which are then in return passed through our Interface to the two VCUs, where the low-level PID-controllers for velocity and steering angle are implemented. They compute the necessary analog voltages for our throttle and braking system and CAN messages for the TKP steering unit. The velocity controller and the steering angle controller have an average tracking error of 0.0614 m/s and 0.1894 degrees, respectively. (see Figure \ref{fig:low-level}) 
	
\begin{figure}[htbp!]
\begin{centering}
\includegraphics[width=0.45\textwidth]{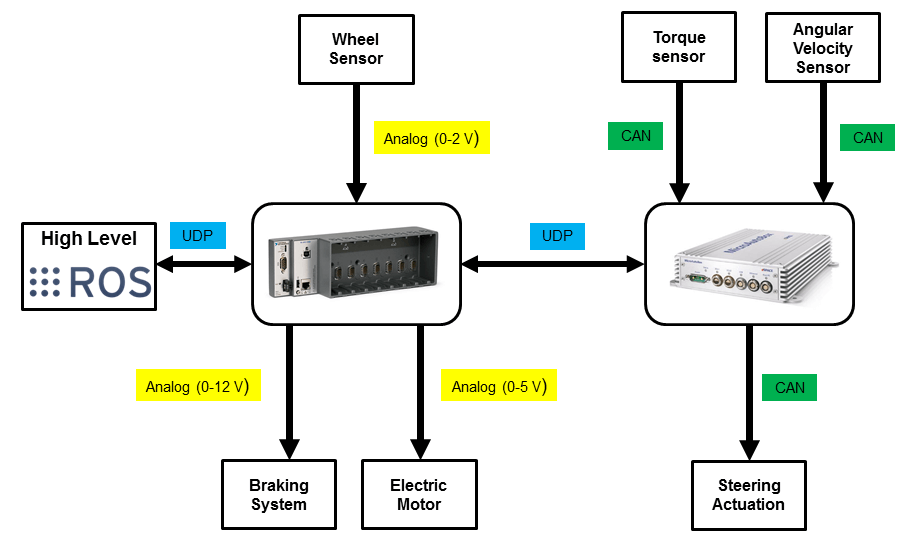}
\caption{Control System layout}
\label{fig:vcu_architecture_final}
\end{centering}
\end{figure}

The information passing between the computer and the two VCUs is established through an interface which we designed (see Figure \ref{fig:vcu_architecture_final}), running at 200 Hz and using the User Datagram Protocol (UDP). The interface was kept as easy as possible to ensure transparency and low latency. Since in this particular architecture the compact RIO is capable of controlling the car's velocity, it is the main VCU and takes control of the steering, permanently checks all UDP connections and inherits all the safety measures on the VCU level.    

\section{System Architecture} \label{sec:system-architecture}
\begin{figure}[htbp!]
	\centering
		\includegraphics[width=0.45\textwidth]{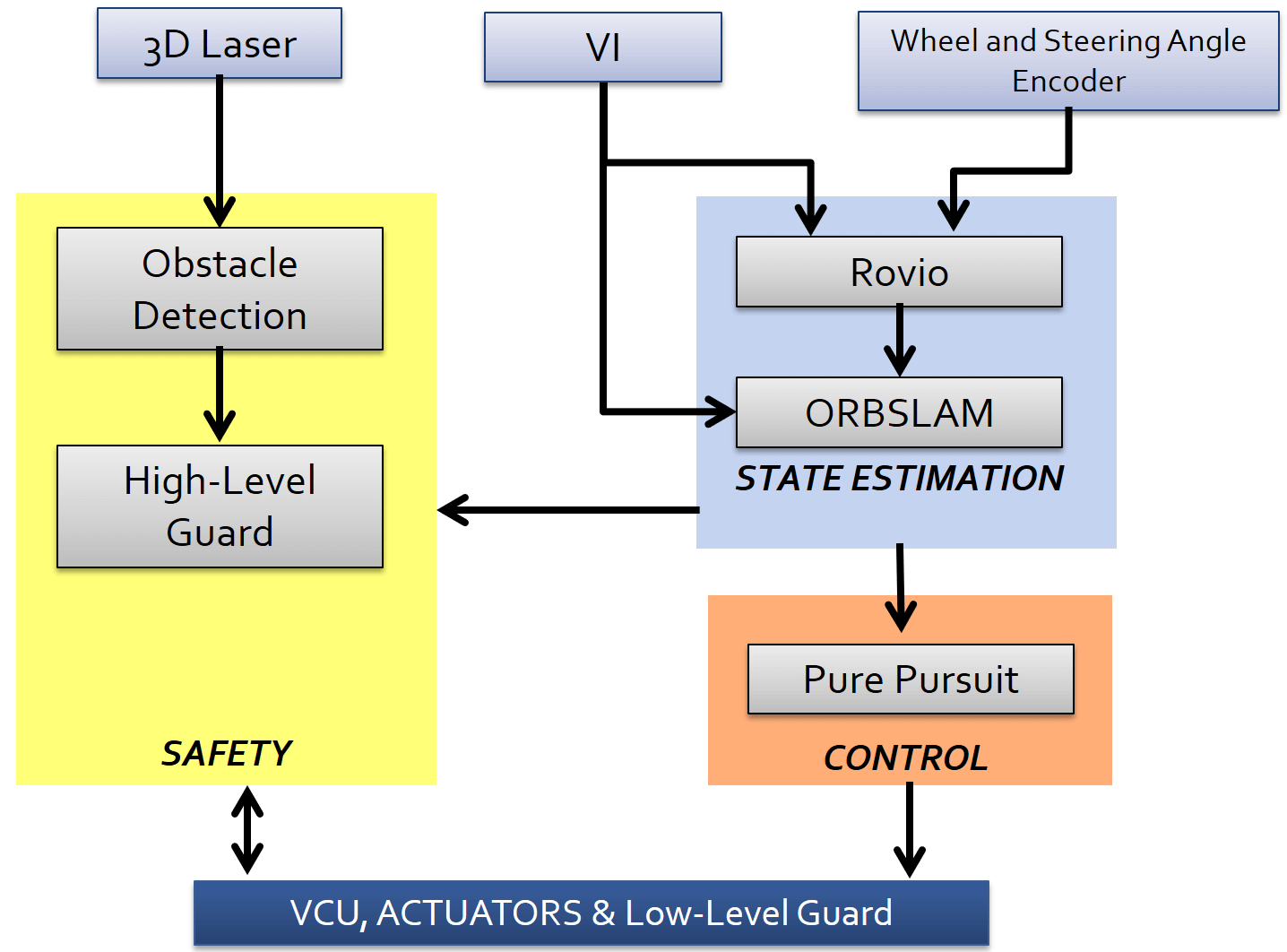}
        \caption{Software Systems layout}
        \label{fig:SoftwareOverview}
\end{figure}

In order to make a vehicle drive autonomously, various different modules are required. 
Figure \ref{fig:SoftwareOverview} provides and overview of sensors and modules used in our pipeline.


\subsection{Teach-and-Repeat}
Our work is based on the teach-and-repeat method. Similar to how localization is performed on a street map, our car localizes using a previously recorded map of visual features. The mapping (teach phase) is essential since there is no global feature map which we could use. In the second step (repeat phase) the car is able to drive the same (or an optimized) path autonomously (repeating). While optimizing the path would be a further development, we follow the same path of the teaching part.

\subsection{State Estimation}
In our system, there are three inputs for estimating the odometry / state of the vehicle: An estimate through vision (optical flow), the IMU, and the vehicle model. 
For the state estimation, the three different kinds of inputs need to be fused.

\begin{figure}[htbp!]
\centering
		\includegraphics[width=0.4\textwidth]{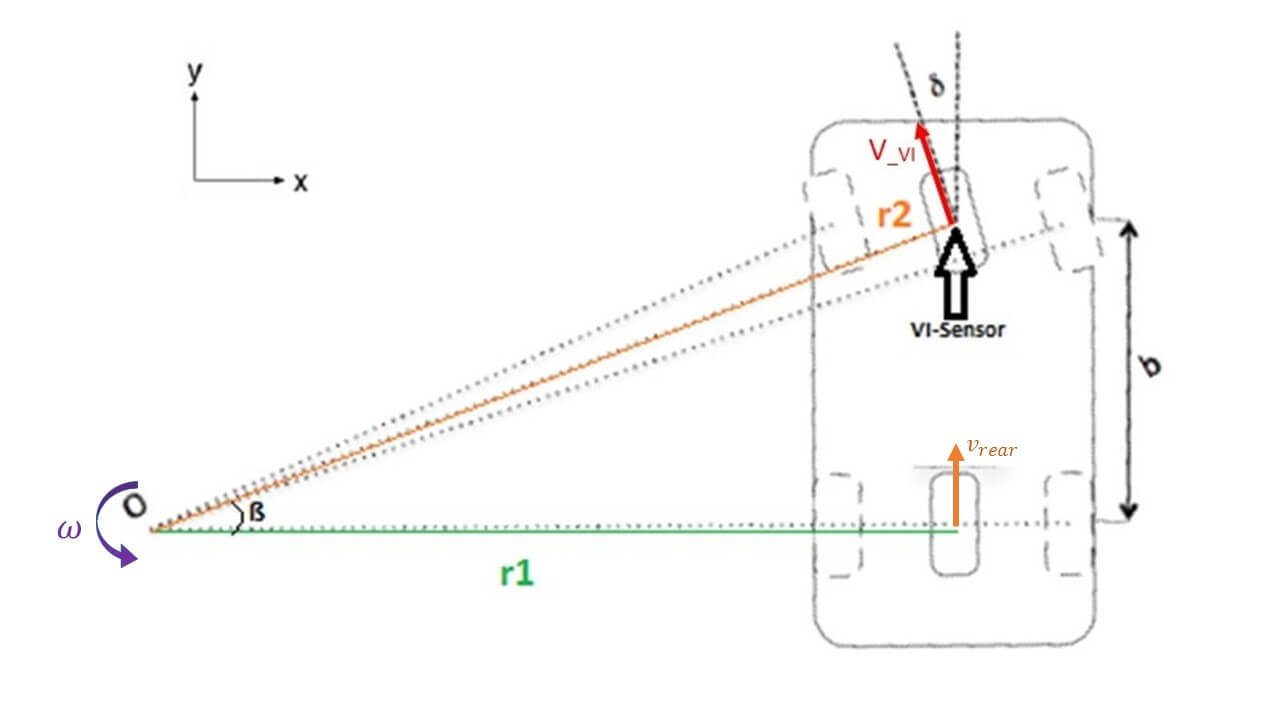}
        \caption{Vehicle Model}
        \label{fig:car_model_final}
\end{figure}

Using the current velocities of the rear wheels as well as the current steering angle in combination with a kinematic vehicle model (see Figure \ref{fig:car_model_final}) one can compute a local estimate of the current velocity in a single plane. Since the car will not be operated at its limits of handling (due to limitations imposed by the state estimation) we decided to use a kinematic bicycle model and make a no slip assumption. An important point is that the velocities have to be expressed in the right coordinate frame which is in our case at the position of the VI-sensor, right between the front wheels. 

\subsection{ROVIO}
In the three dimensional surrounding of a mountain pass it is not sufficient to assume that the plane will be maintained from the start until the end of the drive, therefore we need additional software to compute the three dimensional car odometry. ROVIO is a visual odometry program developed by Michael Bloesch \cite{rovio}. It combines a vision-based odometry estimation with an extended Kalman filter to weight and merge the car model, the vision-based estimate and the IMU measurements. 

Before adding the vehicle model to ROVIO, the odometry estimate was not reliable. The reason for this lay in error-prone IMU measurements in our special application, lacking strong and frequent accelerations compared to a flying UAV but adapting the covariances of the vehicle model and the estimates by the IMU solved the issue.   

Unfortunately ROVIO only uses one CPU core and is therefore limited in computing power. When tracking too many features in a difficult environment (i.e. the majority of features too far away or a grave change in tracked features between consecutive images) the computational load is too large and processing one image takes longer than fetching a new one, resulting in an increasing lag. Lag would lead to a delay in control which can yield vehicle instabilities and therefore has to be avoided. The first approach, reducing the queue size of the incoming image stream, did not help since ROVIO requires a pixel flow. However decreasing the maximum number of features being tracked and the patch size from six to four helped without sacrificing too much accuracy.

Furthermore, another measure to use the limited computing power more efficiently is to discard bad features more quickly and not to waste too much computing power on them. This was achieved by reducing the detection threshold for new features and by increasing the removal factor, if there are not enough free features. There exists a trade-off between a low detection threshold and the maximum number of features. A low maximum number of features needs a higher detection threshold. If not, the set of features changes too fast resulting in a bad estimate. We achieved the best results using a maximum of 20 features and a detection threshold of 0.3. 

\begin{figure}[htbp!]
\centering
  \includegraphics[width=0.45\textwidth]{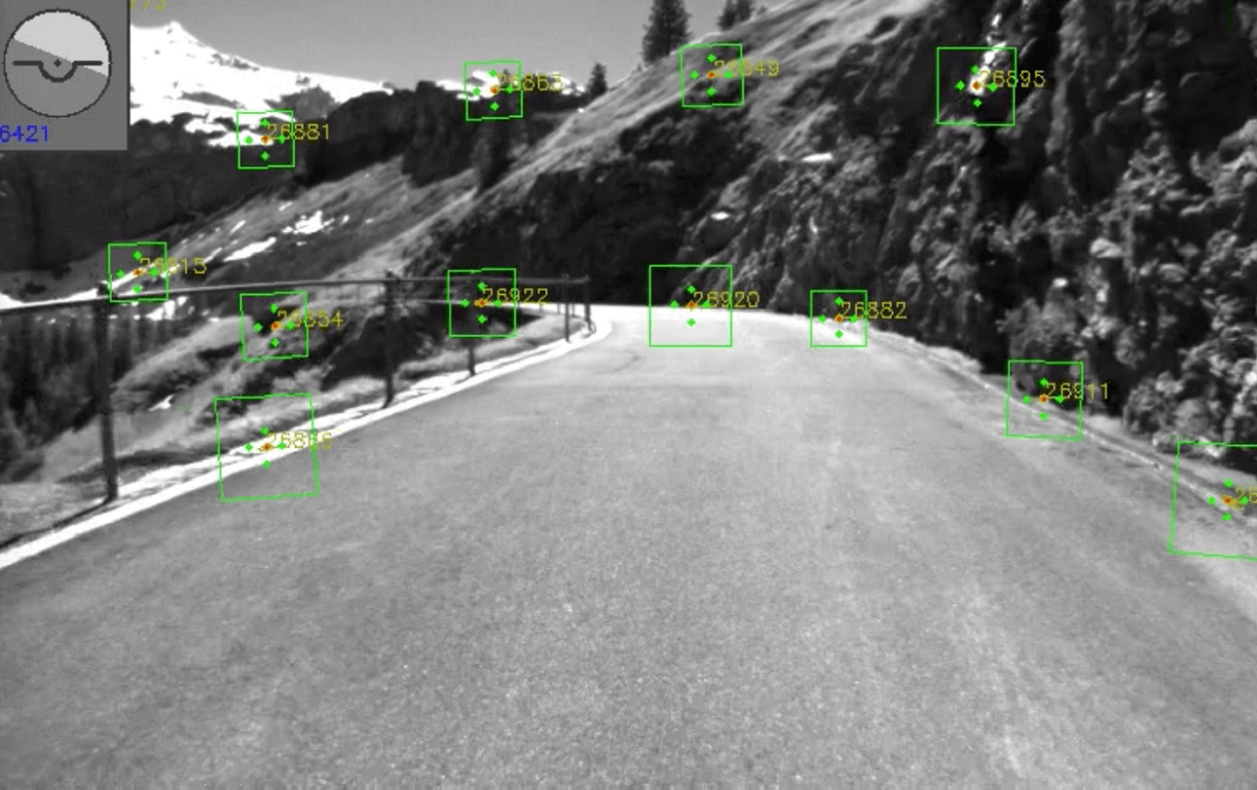}
  
  \vspace{0.1cm}
  
  \includegraphics[width=0.45\textwidth]{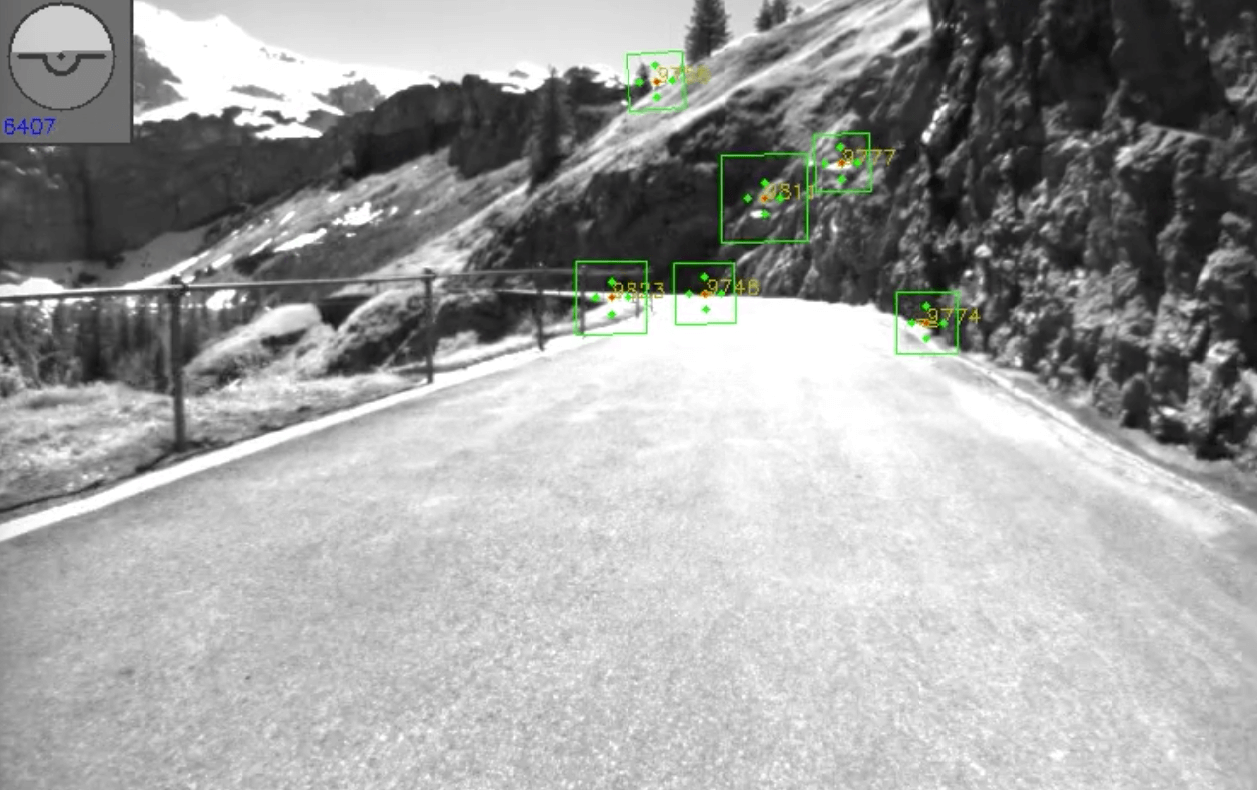}
\caption{Rovio is a visual odometry system, performing optical flow to estimate vehicle geometry independent of environment. Top: Rovio before optimization Bottom: Rovio after optimization}
\label{fig:rovio_principle}
\end{figure}

The most important parameter was the \textit{PenaltyDistance}. Simply put, this parameter determines how homogeneously distributed and close the tracked features can be to each other. Since the close features show a greater change in position between consecutive frames, resulting in an easier velocity estimate, and since sometimes half of the picture can be made up of blue sky without any gradient, we allowed the features to be closer to each other. Nevertheless, a too small \textit{PenaltyDistance} does not help either because this can lead to a few places where all features are clustered, which increases the probability of losing them altogether when the next image arrives. This change had a huge impact and helped to eradicate lag completely throughout the whole climb in conditions like we faced on the Klausenpass. In the future, a further tuning goal could be pushing the limit in terms of velocity, since our finally tuned version had problems when driving through critical environments, where subsequent image frames differ drastically, at velocities greater than 30km/h. A representation of the optimized image for optical flow is shown in \ref{fig:rovio_principle}.

\subsection{ORB SLAM2}
The advantage of a mapping algorithm over a solely visual odometry-based pose estimation can be found in the reduced drift. In terms of localization and mapping we decided to use a vision-based approach (ORB-SLAM2 \cite{orbslam2}) instead of a LiDAR-based one (SegMatch) or a fusion of both (Google Cartographer \cite{hess}). Google’s cartographer was not chosen since it is designed to be used in indoor environments at very low speeds. The reason for preferring ORB-SLAM2 rather than SegMatch lies within our target surroundings and conditions. While with visual teach and repeat, assuming enough light, one can get hundreds of features in every image, a LiDAR-based approach only works when enclosed, geometrical shapes can be detected and extracted which is not the case on a mountain road. 

\begin{figure}[htbp!]
\centering
  \includegraphics[width=0.49\textwidth]{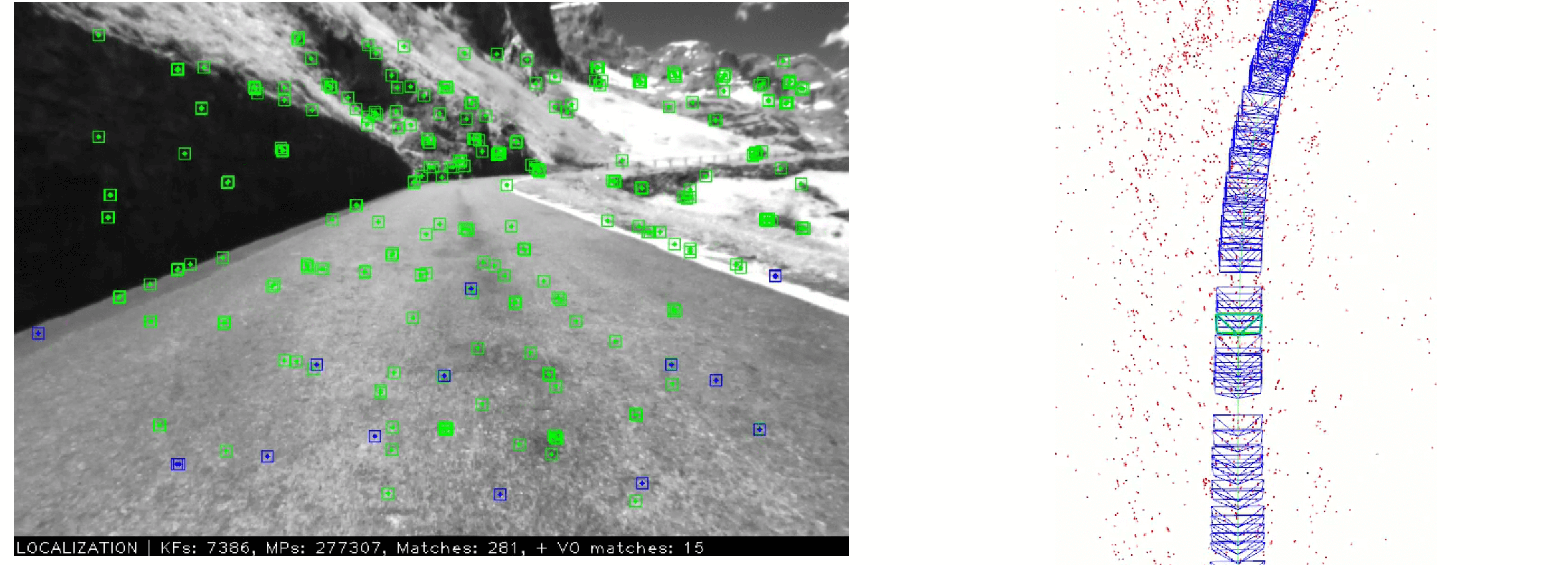}
  \caption{Orbslam operational principle. Green features are known from the teach part. Blue are features that are unknown.}
  \label{fig:Orbslam}
\end{figure}

ORB-SLAM2, developed by Raul Mur-Artal \cite{orbslam2}, uses the raw stereo camera data of the VI as input allowing the calcualtion of each features' pose via triangulation. During teach, ORB-SLAM2 detects features according to the ORB-classification and stores the best suitable features in a global map. An example of such a built map can be seen in Figure \ref{fig:Orbslam}. During the repeat part, ORB-SLAM2 compares the currently seen features with those stored in the map.If there are enough matching features, a pose estimation of the car can be calculated. 

Ueli Graf fused ROVIO and ORB-SLAM2 (\cite{rslam}) which reduces the needed computation power, since ROVIO's odometry estimate is more accurate compared to ORB SLAM's standard internal odometry estimate. Furthermore, in order to reduce small jumps produced by the internal ORB SLAM2 pose optimizer, we added a pose estimate based on the robust and locally accurate odometry input. This also allows a dead reckoning mode which contributes to robustness in the repeating part. The dead reckoning mode occurs when ORB-SLAM2 does not find enough matching features to localize. Then, the pose is calculated by only integrating the odometry until ORB-SLAM2 can localize again in the recorded map. 

Moreover, an algorithm for concatenating several maps, where each map is at a maximum length of 3.3 km, was implemented in order to test long runs without the necessity to load a large map when restarting the system (2min per km up to a length of 3km). This ensures to test more efficiently, reduces the usage of RAM and consequently makes the whole setup more robust. When arriving at the end of a small sub-map, the car stops, the algorithm closes and restarts ORB-SLAM2, loads the successive map, waits until the new map is loaded and then the car itself will continue the autonomous run. 

In order to control the steering angle and the velocity of the car in the repeating phase, a reference path from the teaching part is necessary. The reference path is created by storing the vehicle pose estimates during the teach phase. While driving in the repeat phase, the current pose together with the closest point on the reference path and the odometry estimate are sent to the path following controller.

\begin{figure}[htbp!]
\centering
  \includegraphics[width=0.45\textwidth]{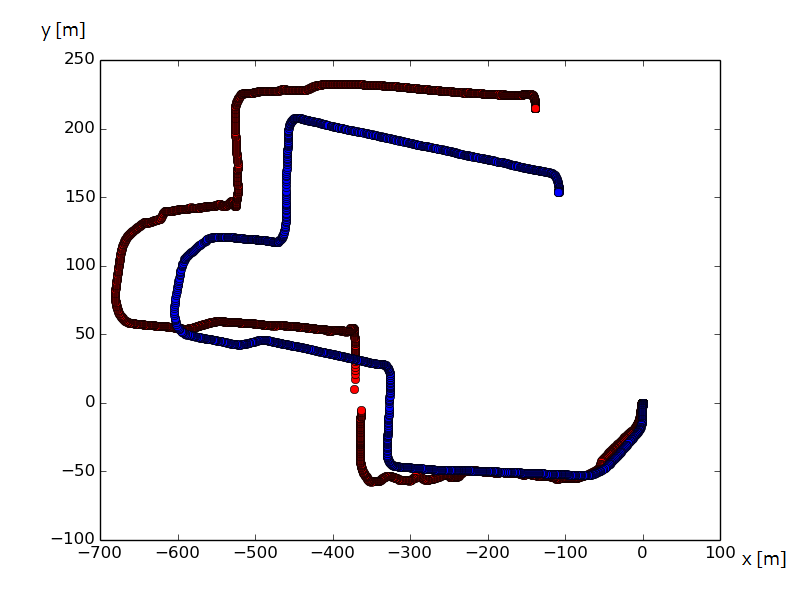}
  \caption{Orslam performance}
  \label{fig:OrbslamPerformance}
\end{figure}

The position estimation of the improved and optimized ORB-SLAM2 can be seen in Figure \ref{fig:OrbslamPerformance}. The dark red reference path is created with GNSS data. The blue path from ORB-SLAM2 has some global drift compared to the GNSS data.  Nevertheless, considering local accuracy, ORB-SLAM2 performed more robust and better than the GNSS pose. In conclusion, our fully optimized ORB-SLAM2 version provides a stable and robust mapping algorithm, which is capable of performing in several different surroundings: Urban areas, rural areas and on mountain roads.

\subsection{Trajectory Following}
In order to follow the recorded path during the repeating phase, appropriate target velocities and steering angles have to be computed which will then be put in practice by the VCUs and actuators. 
The implemented solutions were strongly influenced by our project's specific needs; tracking a recorded path on a curvy and narrow mountain road. Control theory provides innumerable approaches for trajectory tracking, starting from a PID controller over to a simpler geometrical controller, in case of Single Input Single Output (SISO) analysis, up to optimal control approaches such as MPC and LQR for considering a Multiple Input Multiple Output (MIMO) system. With the intention to keep the controllers clearly structured, steering angle and velocity control were decoupled and can hence be resolved separately using SISO approaches.

\subsubsection*{Steering Angle Control}
\begin{figure}[htbp!]
\centering
    \includegraphics[width=0.45\textwidth]{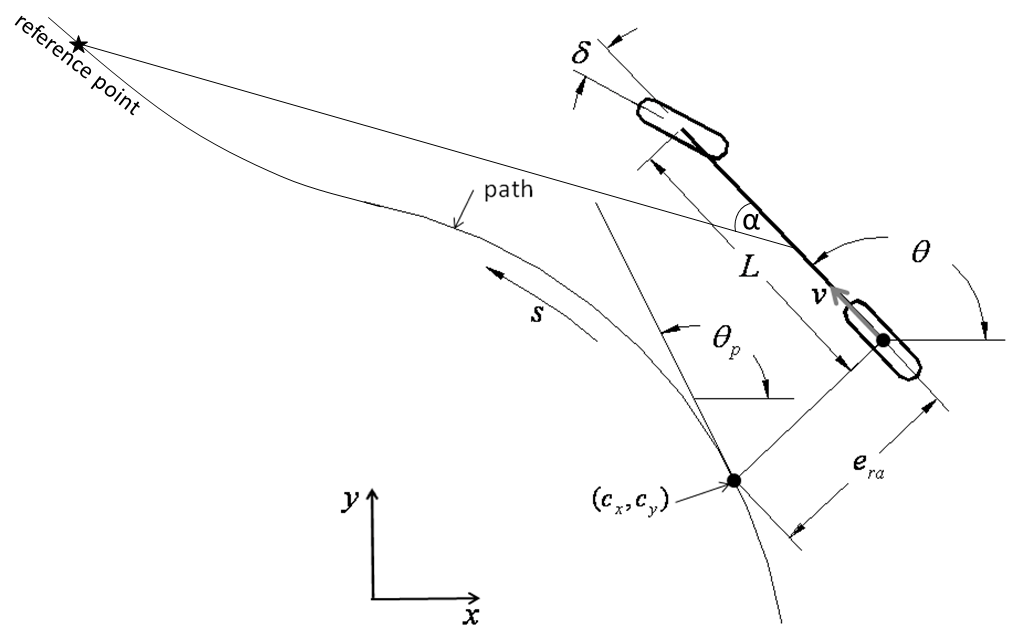}
    \caption{Pure pursuit controller.}
    \label{fig:pure_pursuit}
\end{figure}

The task of the steering angle controller is to compute a target angle from the reference path and the current pose of the car. Since path planning could not be implemented due to missing information about the lane boundaries, the only goal was to track the teaching path as accurately as possible. Inspired by \cite{snider}, we chose to implement the Pure Pursuit Controller: A controller based on a geometric bicycle model. According to \cite{ieee}, at lower speeds, the kinematic bicycle model approximation using simple geometrical relations is at least as accurate as a dynamic model, which estimates the dynamics of the car using the attacking forces and torques. Hence, the moderate velocities in our target environment justified the use of the geometric bicycle model.
Figure \ref{fig:pure_pursuit} illustrates the base concept of the Pure Pursuit steering controller.

The advantages of the chosen path tracking algorithm are its explicit geometrical concept and its independence from velocity control. The controller consists of one explicit equation, allowing the use of non-linearized models like the bicycle model. This spares us the effort of linearization in every single iteration.

\begin{equation}
\delta = 0.8\cdot tan^{-1}(\frac{2Lsin(\alpha)}{l_d})
\end{equation}

	Where:
\begin{itemize}
    \item $\delta$: The required steering angle for tracking the path
    \item $L$: The car's wheelbase
    \item $\alpha$: The angle between the reference point and the local x-axis of the vehicle
    \item $l_d$: The distance of the reference point on the reference path from the vehicle
\end{itemize}

The shown formula computes a steering angle, given a reference point on the reference path we want to reach. The factor $0.8$ was added to the conventional formula due to better performance during test optimization.
In our implementation the look-ahead distance $l_d$ determines the reference point by specifying its distance from the current position and is described by a function, which is proportional to the current velocity with a static ($k2_{lad\_s}$) and a dynamic tuning knob ($k1_{lad\_s}$).

\begin{equation}
l_d = k2_{lad\_s} + k1_{lad\_s} \cdot v_{abs}
\end{equation}

\begin{figure}[htbp!]
\centering
\includegraphics[width=0.45\textwidth]{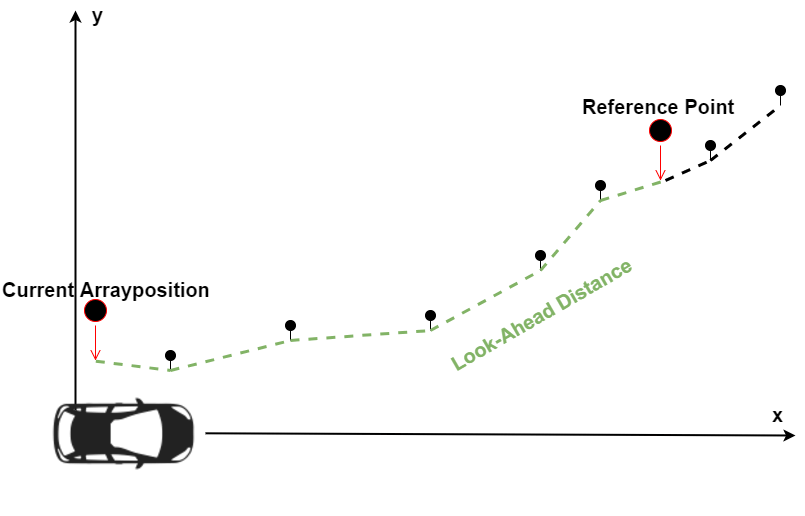}
\caption{Distance of reference points}
\label{fig:SummingUpPathPoints}
\end{figure}

As Figure \ref{fig:SummingUpPathPoints} shows, we decided to use the look-ahead distance ($l_d$) not as an absolute distance but as an Euclidean distance summed up along path points, starting from the point on the path which is currently the closest to the car in order to prevent the car from cutting the curves. 

On the one hand, the look-ahead distance $l_d$ needs a lower bound, 3.5 meters, in order to prevent the car from choosing too close a reference point at low speeds, which causes the car to oscillate around the reference path. On the other hand, an upper bound of 13 meters improves the tracking quality by preventing the cutting of curves. Once an adequate reference point is chosen, it is transformed to the local coordinate frame of the car and then the required steering angle is found by using the Pure Pursuit Controller formula.

\subsubsection*{Velocity Control}

The eRod’s velocity is the second control variable. To get an easily tunable reference velocity, a connection to the velocity of the teaching part is kept, while independently calculating a feasible velocity with respect to the upcoming path section. To determine a physically meaningful reference velocity we apply Coulomb's friction law between tyre and asphalt:

\begin{equation}
v_{ref\_physical} = \sqrt{\mu_{fric}\cdot g_{earth}\cdot curve_{radius}}
\end{equation}

Where:
\begin{itemize}
    \item $\mu_{fric}$: The static friction coefficient between the tyres and the asphalt
    \item $g_{earth}$: The earth's gravitational pull
    \item $curve_{radius}$: An approximated curve radius
\end{itemize}

This curve radius is found by fitting circles in the region of interest. The region of interest is determined by a specific distance which is calculated similarly to $l_d$ in the Pure Pursuit Steering Controller:

\begin{equation}
l_d = k2_{lad\_v} + k1_{lad\_v} \cdot v_{abs}
\end{equation}

$k1_{lad\_v}$ and $k2_{lad\_v}$ represent an empirically determined dynamic gain and a static offset. Since the physical limit reaches infinity on an approximately straight path, the velocity is constrained to be lower than two upper boundaries. One is related to the velocity at the current position during the teaching part ($v_{teach} + v_{freedom}$), where $v_{freedom}$ is a user-chosen parameter. The other is a maximum velocity based on the given environment ($max_{abs\_vel}$) and also defined by the user. After this first step, the reference velocity is:

\begin{equation}
	v_{ref}=min ( v_{ref\_physical} ; v_{teach} + v_{freedom} ; max_{abs\_vel} )
\end{equation}

\begin{figure}[htbp!]
\centering
   \includegraphics[width=0.49\textwidth]{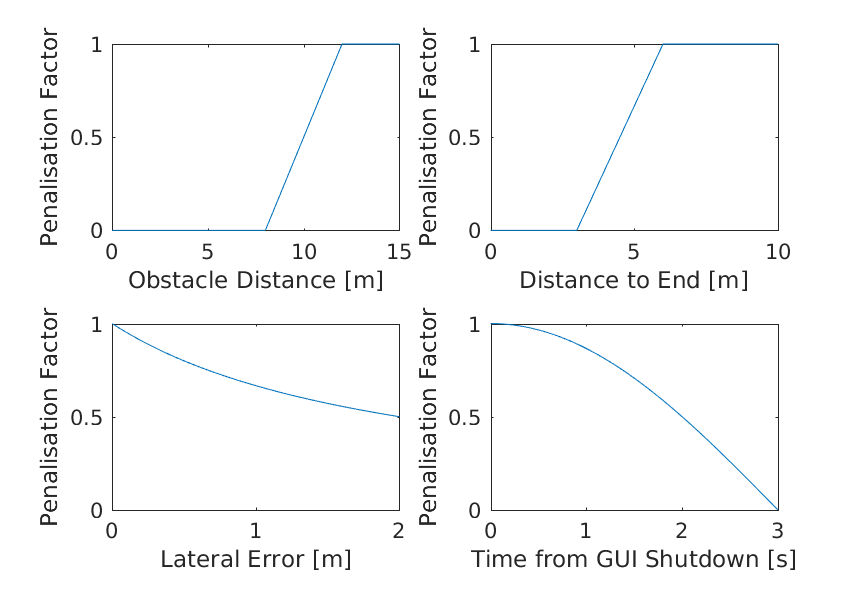}
   \caption{Our four different penalization factors.}
   \label{fig:Penalisations}
\end{figure}

During testing, we noticed that the velocity is almost always bounded by the teaching or user-chosen velocity limit and not by Coulomb's friction law.
In the next step, the previous reference velocity $v_{ref}$ is further adapted by applying different penalizations which in some cases might even shrink $v_{ref}$ to zero. This is done by multiplying $v_{ref}$ with four different factors. The four factors are the penalization factor for a certain obstacle distance, lateral error, distance to the end of our path and time from GUI Shut-down. Each of the functions, shown in Figure \ref{fig:Penalisations}, determine the penalization factor and can be easily adapted.

\section{Safety Systems} \label{sec:safety}
\begin{figure}[htbp!]
  \centering
  \includegraphics[width=0.3\textwidth]{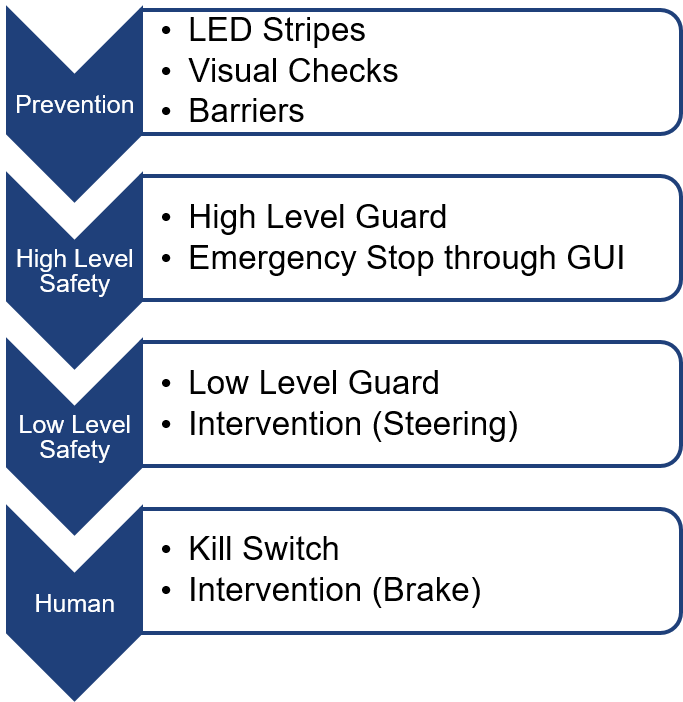}
  \caption{Cascading security levels}
  \label{fig:security_levels}
\end{figure}

When developing an autonomous vehicle, the most important goal is to guarantee a maximum amount of safety at any point in time. Therefore we set up four safety layers (see Figure \ref{fig:security_levels}) and decided to always have a person in the driver seat who can take over the eRod at any time.

\subsubsection*{Obstacle Detection}
The ability to react to obstacles blocking the upcoming reference path is crucial for the safety of our car and its surrounding, and in theory eliminates the need for a person sitting in the driver seat. The most important characteristic of a reliable obstacle analysis on a curvy mountain road is that the implemented solution takes the future trajectory of the car into account. This allows the car to drive through tight curves without classifying the roadsides as obstacles and works as follows: In the first step, all potential obstacles in the car's proximity must be detected. In the second step, the grid map is analyzed whether any of the obstacles block the upcoming path section.

\begin{figure}[htbp!]
\centering
   \includegraphics[width=0.47\textwidth]{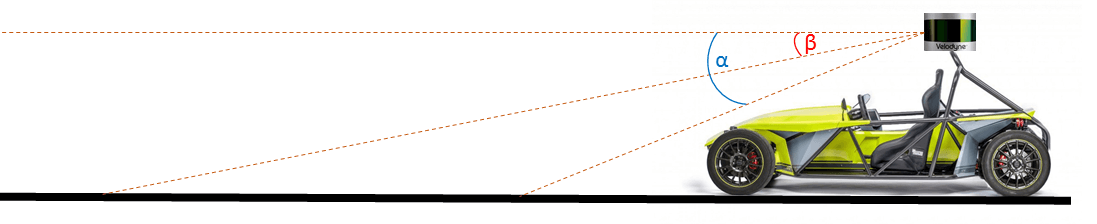} 
   \caption{Visualisation of vertical angles (meaning $\alpha$, $\beta$).}
   \label{fig:LaserBeam}
\end{figure}

\paragraph*{3D Object Detection}
The obstacle detection algorithm relies on the point clouds received from the LiDAR. At first, all points received from the LiDAR are assigned to certain vertical angles (e.g. $\alpha$, $\beta$) as shown in Figure \ref{fig:LaserBeam}. 

\begin{figure}[htbp!]
  \centering
   \includegraphics[width=0.3\textwidth]{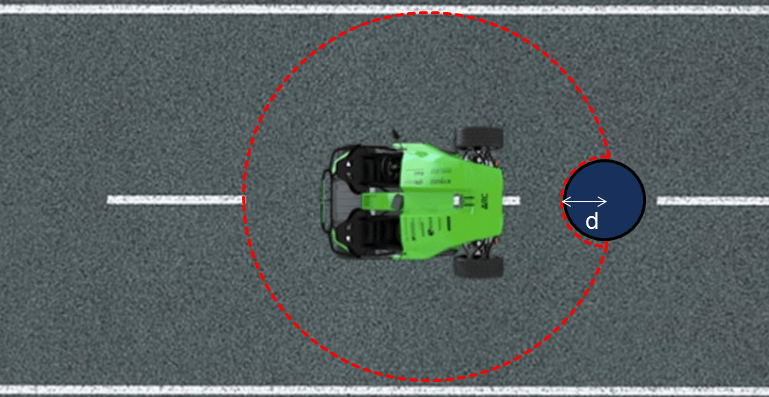} 
   \caption{Schematic of the obstacle detection.}
   \label{fig:ObstacleDetection}
\end{figure}

The algorithm is based on the fact that the points belonging to one vertical angle form a circle around the car (see Figure \ref{fig:ObstacleDetection}) resulting in a constant distance. In case of an ascending road, the points on the road form an ellipse. However, the change of the distance between consecutive points on the ellipse and the laser is small compared to the grave change in distance if an obstacle is present, which results in a simple detection of obstacles. 

For each vertical angle, all points outside a triangle of 90 degrees, which turns in line with the current steering angle, in front of the car are filtered out to analyze the region of the street only. Through statistical analysis of the remaining points, a distance interval can be determined which contains most of the points and therefore, assuming no obstacles blocking the majority of the road, returns the distance between the LiDAR and the road. In the next step, the x- and z- coordinates of adjacent points are compared as follows: If one of the adjacent point was detected as a point on the road and the summation of the difference in x- and z- direction lays in a certain empirically determined tolerance, both points are labeled as road. If however, the difference in distance is larger than the tolerance, the other point is marked as belonging to an obstacle. The y- coordinate is ignored because not each laser beam gets reflected resulting in an incorrect distance in y-direction.   
Furthermore, if there are any points at a vertical angle of 0 degree, they are automatically labeled as obstacles, making it possible to detect a flat wall in front of the car. In the final step, all points labeled as obstacles are projected into the horizontal plane and marked in an occupancy grid map. This grid map is then sent to the grid map analyzer. 

However, our current 3D-object detection algorithm has a decreased ability to detect a flat object blocking the whole road. The positioning of the LiDAR in the front of the car just above ground might result in a simpler, more straight-forward and more precise obstacle detection but we wanted to keep the possibility to implement a 360 degree obstacle detection.
	       
\paragraph*{Grid map Analyzer}
Due to the fact that our system does not include lane tracking, we developed an obstacle detection which stops in front of the obstacle and is not capable of driving around it which is anyways very difficult on narrow roads like the Klausenpass. Because of our modular system, this would still be a possible further development. 
    
\begin{figure}[htbp!]
\centering
   \includegraphics[width=0.1\textwidth]{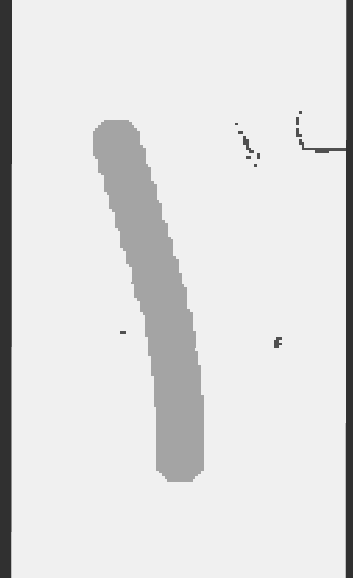} 
   \caption{Rviz visualisation of the final grid map.}
   \label{fig:pathprojection}
\end{figure}

Not all objects displayed in the grid map represent a potential danger for the car. That is why a grid analysis is required which returns the distance to the closest path blocking obstacle. The basis therefore is the estimate of a 'dangerous zone' which at first has the form of the estimated future path. The thickness of this zone is the broadness of the car plus a certain tolerance which is proportionally dependent on the tracking error (i.e. the lateral deviation from the planned path). After having computed the 'dangerous zone' it is added to the grid map. Figure \ref{fig:pathprojection} visualizes the final grid map including the 'dangerous zone' (grey) and the detected obstacles (black). The comparison between the 'dangerous zone' and obstacles reveals if there are any obstacles on the car's future trajectory and if yes, the shortest obstacle distance is returned. 

Depending on the measured distance, a distinction between a critical and non-critical situation is performed algorithmically, while the critical distance interval is a function of the actual velocity: 
$[0;(\frac{v\textsubscript{abs} \cdot 3.6}{10})^{2}]$. 

Critical obstacles trigger an emergency brake while non-critical ones are passed to the controller which decreases the reference velocity through the corresponding penalization factor. If an obstacle causes the car to stop but the car's future trajectory is freed again, the car automatically continues the run and returns to the reference trajectory tracking task. 

If an obstacle is so small that it lies between two LiDAR beams, the obstacle cannot be detected on a frame-by-frame basis. This results in objects appearing and disappearing from the grid map causing the obstacle distance and therefore the reference velocity to vary significantly in short time periods. To prevent this, we added a time delay whenever a recognized obstacle disappears. During this time, the grid map analyzer simply assumes that the object is still present. To manage the case in which the car comes to a stop exactly at a distance where the obstacle cannot be detected, another time interval of three seconds was introduced. During this interval, our car slowly moves forward to check if the obstacle is still present. This delay guarantees a smooth halt in front of small objects that are present but cannot be detected constantly.  

The obstacle distance is sent to our guard algorithm, which is checking this value amongst many others and eventually takes action by sending the emergency brake signal to the cRio and by decreasing the reference velocity to 0 m/s while the steering controller keeps running. The reason for this is that most of the times when an emergency brake will be actuated, it is safer to continue controlling the steering and not to steer randomly or to drive straight. Other safety critical values are for instance the car's orientation, the tracking error, the output of the System's Watchdog and the VCUs' status. 

\subsubsection*{Further Safety Measures}

The VCUs are the most essential pieces of hardware in order to guarantee a safe and robust system because of their direct connection to the actuators. Due to our special architecture, all safety measures are performed by the cRIO because of its capability of controlling the cars' velocity and of changing the visual LED output. 
In case of any internal failure, e.g. an emergency stop sent by the computer, interface breakdown, human steering intervention or the end of an autonomous drive, the velocity controller is disabled and the cRIO immediately switches the LEDs from blue to green. The first three previously mentioned events also trigger the emergency braking procedure which causes a full stop of the vehicle.  

The human steering intervention event is detected whenever the steering torque sensor measures a total torque of more than 7.5 Nm acting on the steering wheel. This threshold ensures that disturbances, e.g. resulting from uneven roads, do not lead to a shut down.

It is important to establish safety measures on the software level, but the best way to guarantee safety is their implementation on the physical hardware level.  

A complete failure of the eRod 12V system will not stop our autonomous system from running because we installed additional batteries which can provide all the electrical power needed by our system for about 15 minutes. 
An emergency button next to the driver seat was installed which physically cuts all actuators from power, and definitely stops the autonomous system from running. 
All of the connecting and disconnecting processes (which also change the color of the LEDs) are executed by industrial relays which guarantee fail safeness for millions of switching events. 

Moreover, the implementation of the braking system allows the driver to brake in every situation and physically override any command from the autonomous vehicle controller and thereby reduce the velocity of the autonomous system. Like this, the driver cannot only over-steer but also over-brake the autonomous system at any time.

We are very proud of having established such a robust, quick and safe overall system together with its hardware implementation. This is one of the main reasons why we never had any accident with our vehicle throughout our 1000 km of testing. It turned out that for the driver, the steering intervention is the best way to stop the system in case of any failure or danger. Furthermore, if the computer wants the cRIO to trigger an emergency brake, this takes less than 20 ms.

\section{Results} \label{sec:results}

\begin{figure}[htbp!]
\centering
  \includegraphics[width=0.45\textwidth]{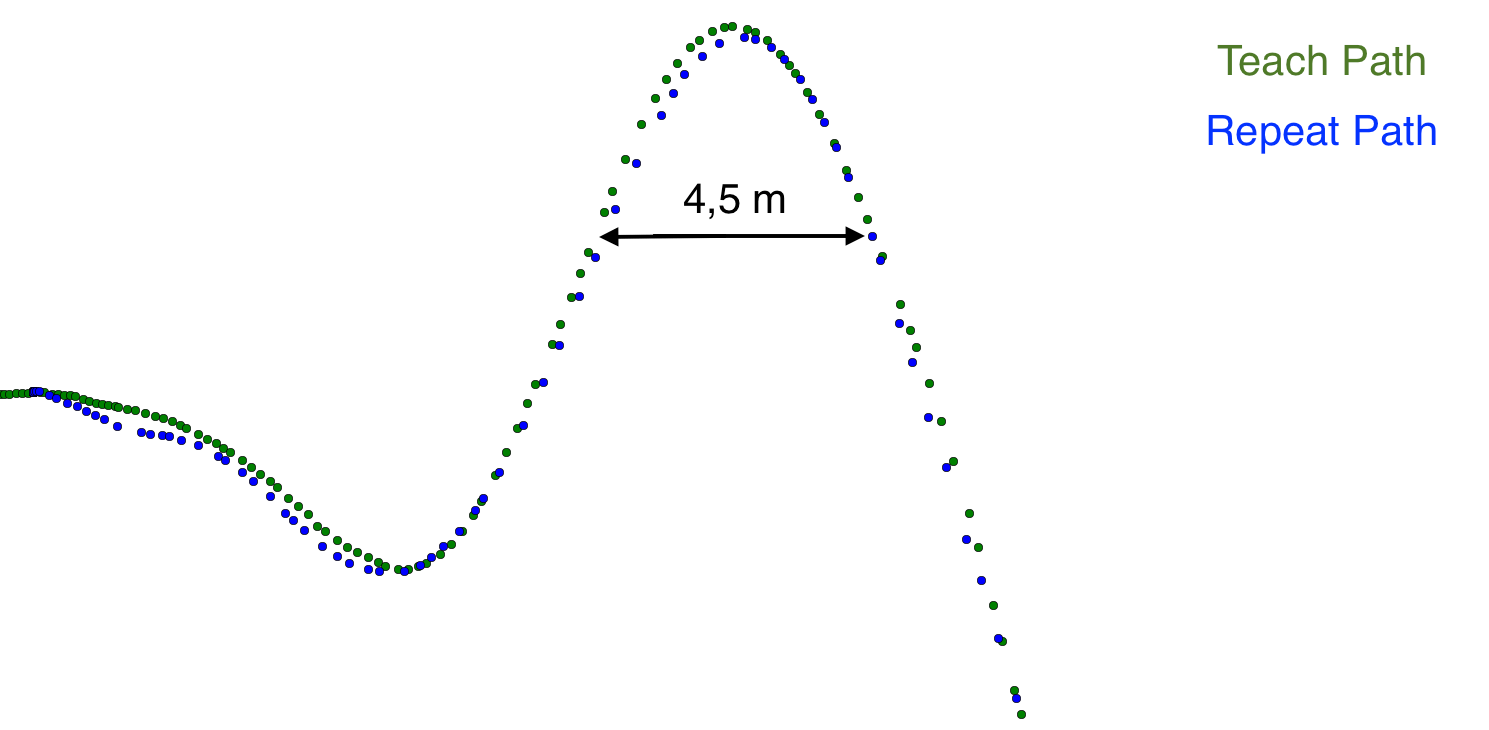}
  \caption{Performance at 20-25 km/h}
  \label{fig:middle_edited}
\end{figure}

\begin{figure}[htbp!]
\centering
  \includegraphics[width=0.45\textwidth]{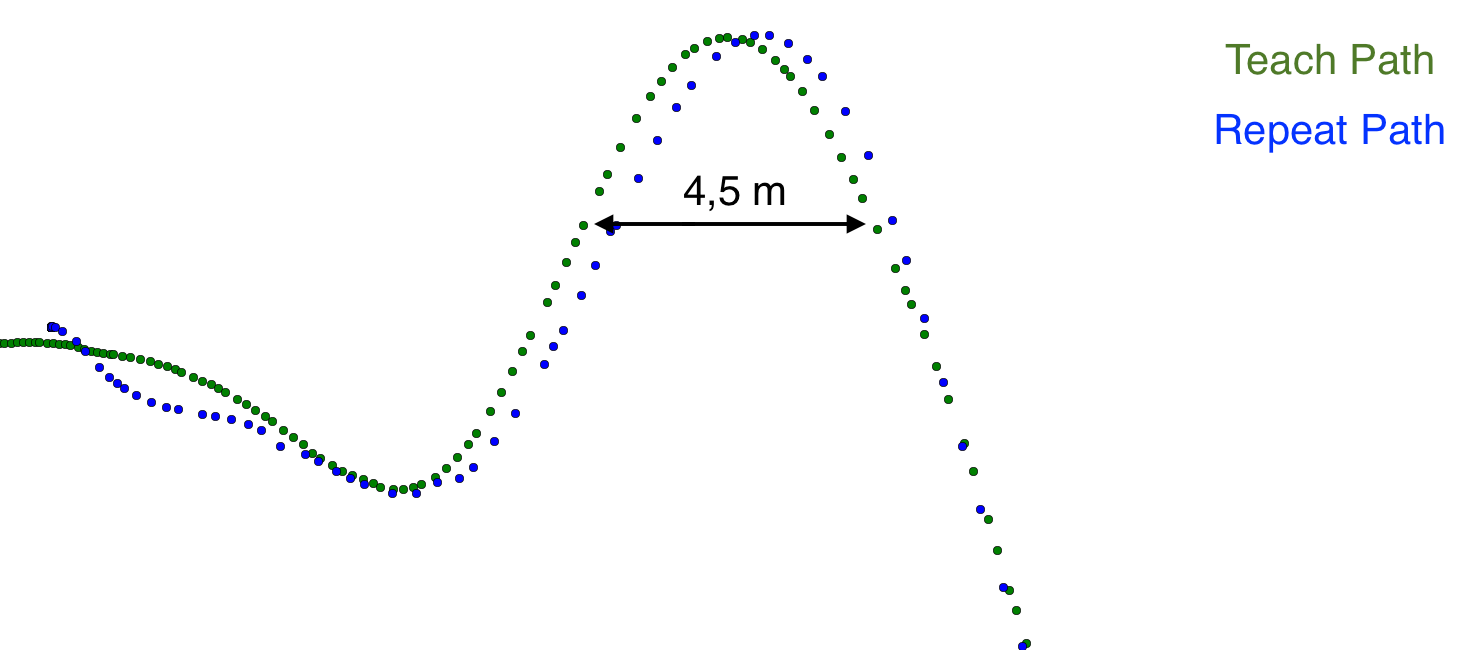}
  \caption{Performance at 25-30 km/h}
  \label{fig:fast_edited}
\end{figure}

The overall system demonstrates a good teaching and path following performance without being too aggressive (see Figure \ref{fig:middle_edited}). At velocities up to 25 km/h, our system's median path tracking error is about 10 cm. 
As it can be seen in Figure \ref{fig:fast_edited}, the curve cutting behavior emerges with increasing velocities because of the high-level steering controller. 

Our system was mainly optimized for usage on the Klausenpass, where most of the features are far away from our vehicle. Our system works up to about 35 km/h, but is nevertheless limited to a path length of 3.5 km using a single map only. Due to the limiting testing time, we did not yet have the chance to test longer runs by concatenating several sub-maps (of 3.3 km each). The inclination angle of our stereo camera allows a lateral tracking error of (+/-2m) without losing localization fix. The orientation divergences must not be greater than (+/- 20 degrees). Our system stays in dead reckoning mode after 2 seconds of missing re-localization. Missing localization for longer than this period results in a termination of the autonomous run. Due to our vision-based approach, we are strongly dependent on the light conditions. In most cases we were able to localize in a map for about 2.5 hours before having to do a new teaching / mapping run. Bright light with a low angle of incidence as well as routes taught on overcast days and repeated on sunny days (or the other way around) are particularly problematic.

All in all, our system behaves similarly as Furgale's Mars rover (\cite{furgale}) in many ways, e.g. light dependence. Nevertheless, because of our huge effort in testing and optimization the state estimation of our system seems to be very robust in terms of the repeat deviation and our system works under higher velocities.

\begin{figure}[htbp!]
\begin{centering}
\includegraphics[width=72mm]{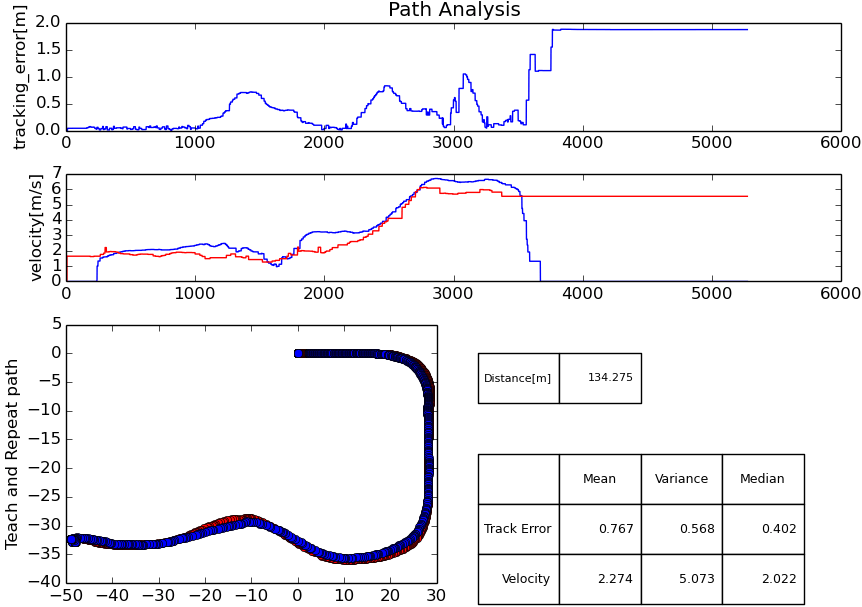}
\caption{Logging file displayed during each teach/repeat phase}
\label{fig:logging_file}
\end{centering}
\end{figure}

\section{Conclusion} \label{sec:conclusion}

In this paper we outline the results of modifying a Kyburz eRod into a fully autonomous platform. An exhaustive description of the hardware and systems was provided, as well as detailed test results that show our state estimation and path following control under tough real-world conditions.

Our vision was to conquer the Klausenpass with a vision-based teach-and-repeat method only. We are proud that we realized our vision to autonomously drive up this difficult and unforgiving mountain road for more than three kilometers. Using vision-based navigation in such a demanding environment applying the teach and repeat method is an effective showcase for the potential of this technology. 
With further refinements to  deal with changing light conditions and remote features, this technology promises great opportunities for the future of autonomous driving and may also be extended to other applications such as indoor navigation. 

The fact that ten bachelor students built an autonomous vehicle within nine months using a small budget compared to other companies also demonstrates the feasibility of autonomous driving. With our system, we emphasized the safety of autonomy and increased the enthusiasm of the public for this exciting topic. 

For future efforts, merging several maps of the same location recorded under different lighting conditions could even reduce the main disadvantage of visual localization, i.e. its dependence on lighting conditions. As a further improvement, we could create a global map with all the streets we have ever driven. If this global map would include the desired route, the teaching phase would be obsolete. In the near future, maybe the most feasible application would be to implement this technology for the daily travel of the same route, e.g. the way to the workplace of a commuter could be driven autonomously. 

\section*{Acknowledgments} 
The authors would like to thank Hiag Data, Kyburz-Switzerland AG, ETH Zurich and the Autonomous Systems Lab (ASL), Thyssenkrup presta AG, National Instruments, u-blox, Helbling and Hasler for their financial, technical, and moral support, without which this project would not have been possible.

\bibliographystyle{IEEEtran}
\bibliography{evs30abstracttemplate}
\onecolumn
\section*{Authors}

\begin{tabular}{cccc}

\begin{minipage}[b]{21mm}
\includegraphics[width=20mm]{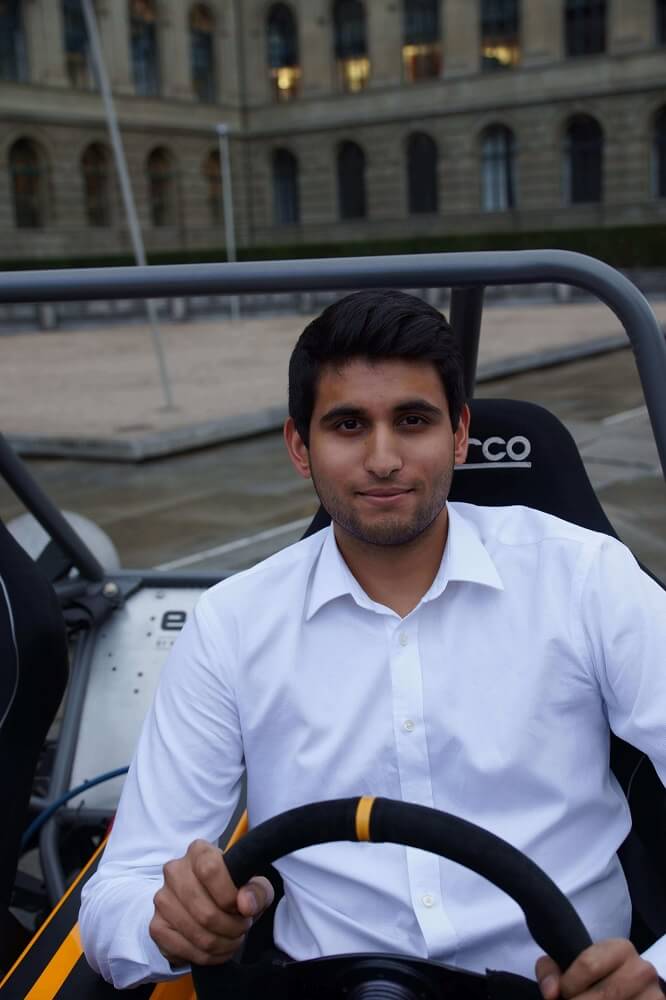}
\end{minipage} 
& 
\begin{minipage}[b]{60mm}
\small
Nikhilesh Alatur currently studies mechanical engineering with a focus on Robotics at ETH Zurich in his final bachelor semester. In the project he worked on vehicle simulation, path following and low level control. 
\end{minipage}
&
\begin{minipage}[b]{21mm}
\includegraphics[width=20mm]{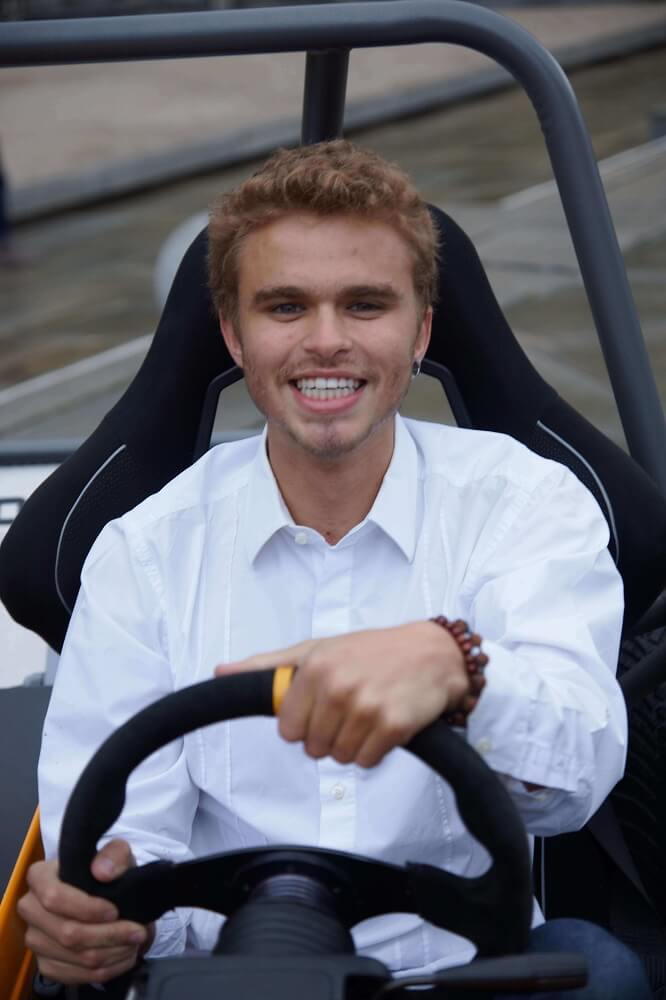}
\end{minipage} 
&
\begin{minipage}[b]{60mm}
\small
Nicholas B{\"u}nger is a final-year bachelor student in mechanical engineering at the ETH Zurich and is responsible for the System Engineering and Telemetry in Project-ARC.
\end{minipage} 
\\
\end{tabular}
\vspace{1cm}

\begin{tabular}{cccc}
\begin{minipage}[b]{21mm}
\includegraphics[width=20mm]{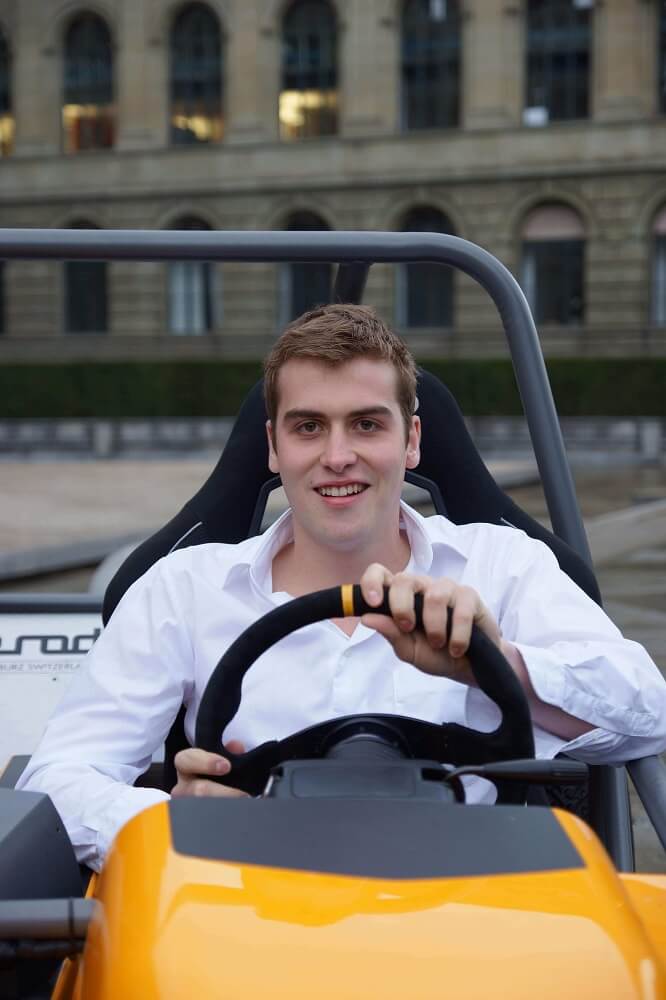}
\end{minipage} 
& 
\begin{minipage}[b]{60mm}
\small
Robin Deuber studies mechanical engineering BSc in his final year at ETH Zurich. His study focus lies on control systems and robotics. In the project ARC he is responsible for control, actuation and safety.
\end{minipage}
&
\begin{minipage}[b]{21mm}
\includegraphics[width=20mm]{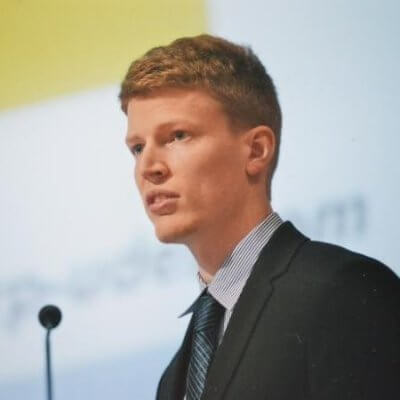}
\end{minipage} 
&
\begin{minipage}[b]{60mm}
\small
Renaud Dub\'e completed his Bachelor and Master at the University of Sherbrooke in Canada. He is currently PhD student at the ETH Zurich and focuses on 3D perception for autonomous vehicles using laser range finders.
\end{minipage} 
\\
\end{tabular}
\vspace{1cm}

\begin{tabular}{cccc}
\begin{minipage}[b]{21mm}
\includegraphics[width=20mm]{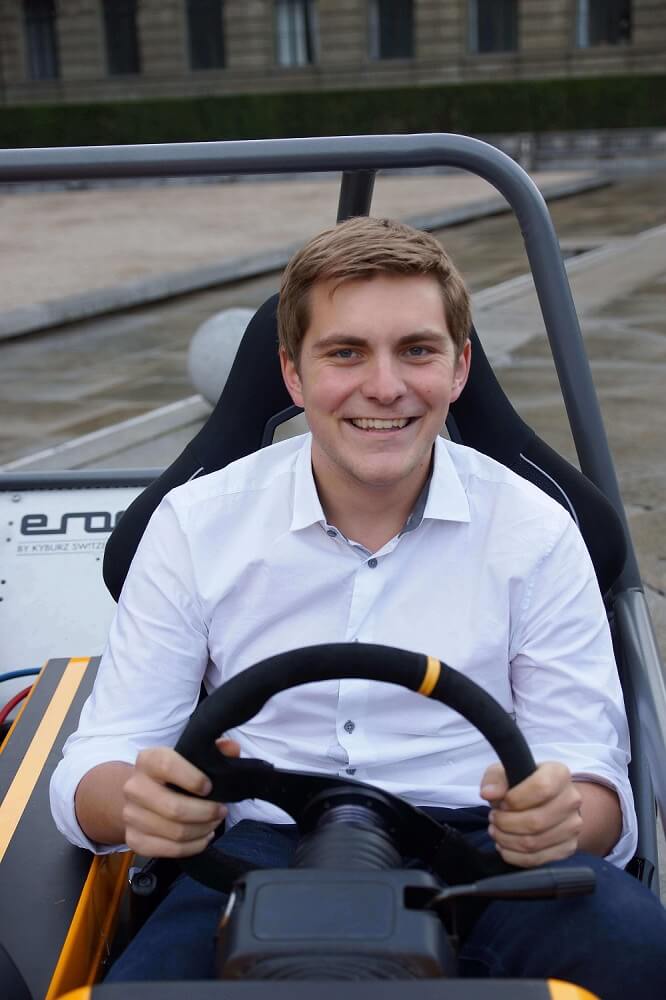}
\end{minipage} 
& 
\begin{minipage}[b]{60mm}
\small
Niklas Funk is a final-year bachelor student in Electrical Engineering at ETH Zurich and will start with the master's programme in Robotics, in fall 2017. He worked on Project-ARC's State Estimation and Interface as well as on the electrical supply of all Sensors, Actuators and Computer Hardware.
\end{minipage}
&
\begin{minipage}[b]{21mm}
\includegraphics[width=20mm]{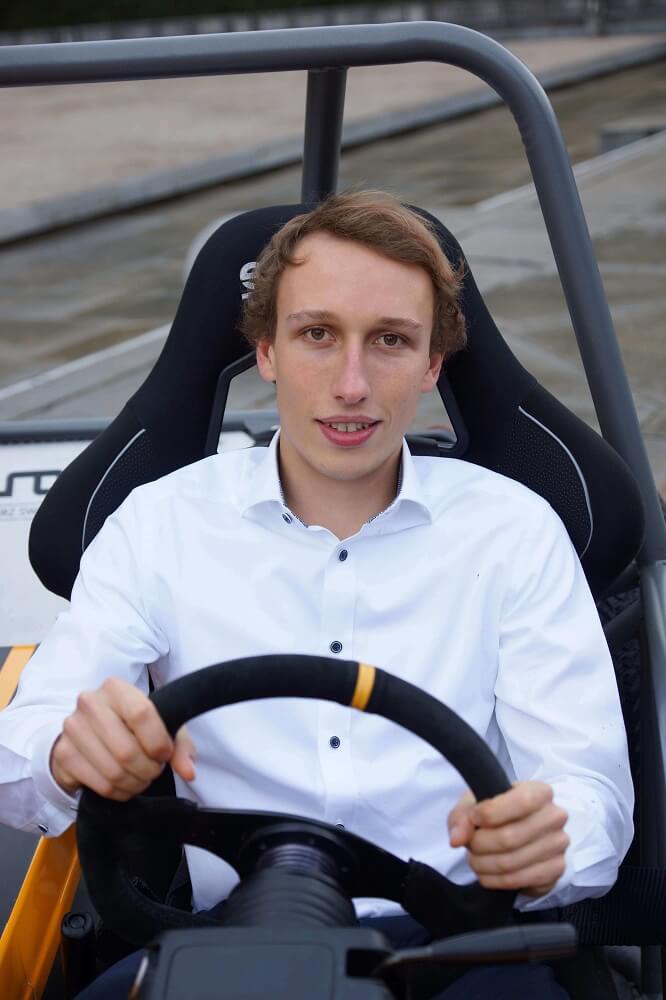}
\end{minipage} 
&
\begin{minipage}[b]{60mm}
\small
Frederick Gonon studies mechanical engineering in the 6th semester at ETH Zurich. He was responsible for the electric steering \& braking system and for the construction of the sensor mount as well as the low-level velocity control. In his bachelor thesis he developed a dynamic vehicle model for a launch control to maximise the acceleration and stability of the car. 
\end{minipage} 
\\
\end{tabular}
\vspace{1cm}

\begin{tabular}{cccc}
\begin{minipage}[b]{21mm}
\includegraphics[width=20mm]{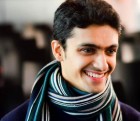}
\end{minipage} 
& 
\begin{minipage}[b]{60mm}
\small
Raghav Khanna is a PhD student in Robotics at ETH Zurich and focuses on perception and mapping for autonomous aerial vehicles using visible and multi-spectral cameras.
\end{minipage}
&
\begin{minipage}[b]{21mm}
\includegraphics[width=20mm]{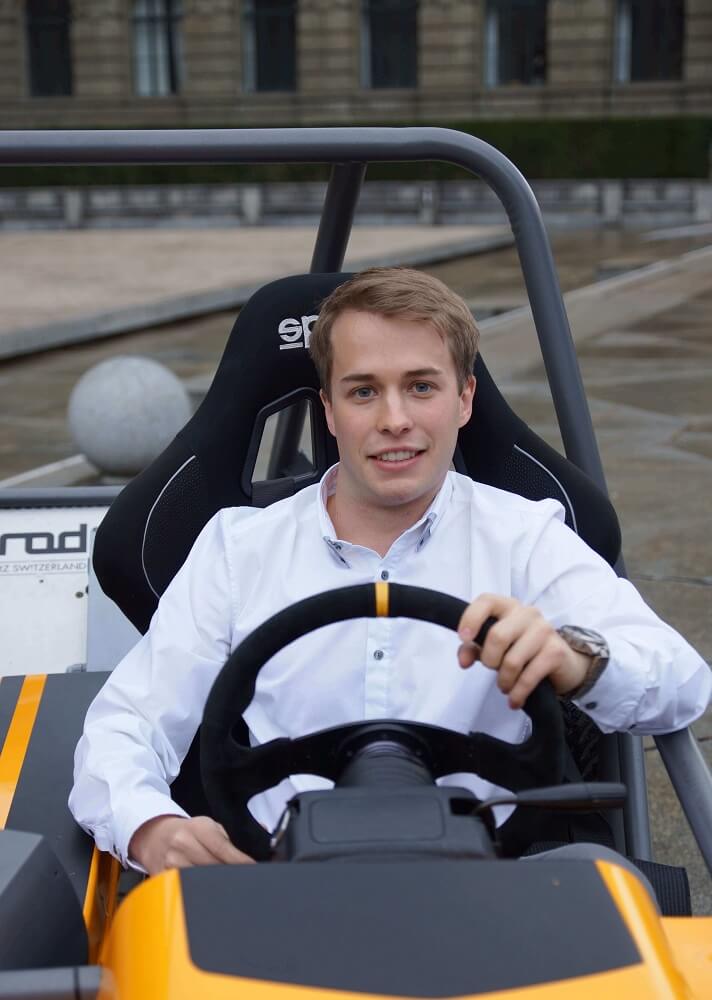}
\end{minipage} 
&
\begin{minipage}[b]{60mm}
\small
Nico Messikommer studies mechanical engineering in the 6th semester at ETH Zurich. During Project ARC, he worked on the implementation of a localisation \& mapping based state estimation as well as the development of an obstacle detection.
\end{minipage} 
\\
\end{tabular}
\vspace{1cm}

\begin{tabular}{cccc}
\begin{minipage}[b]{21mm}
\includegraphics[width=20mm]{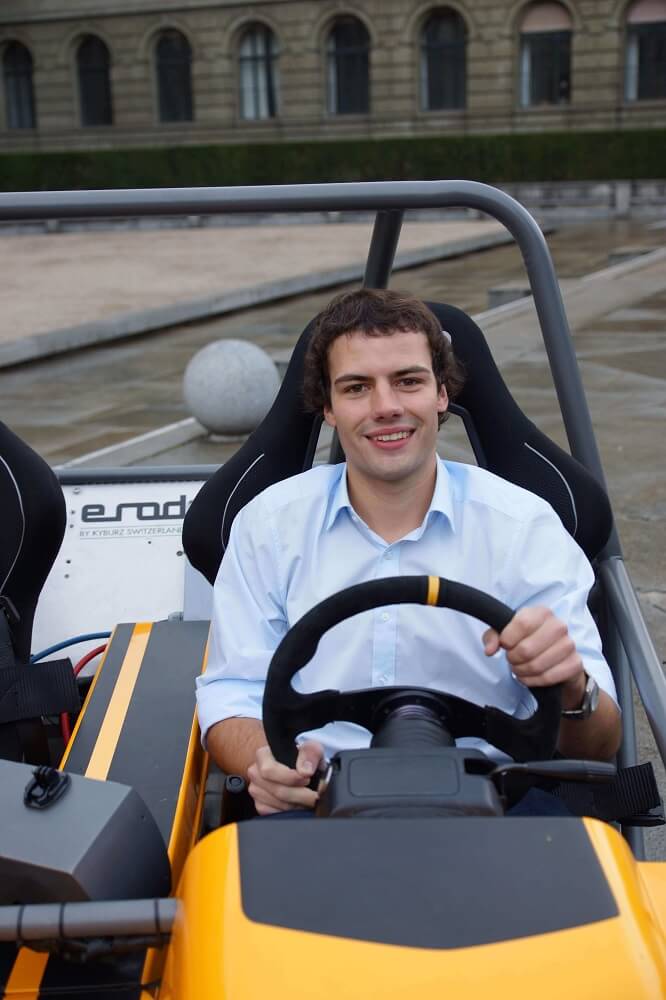}
\end{minipage} 
& 
\begin{minipage}[b]{60mm}
\small
Julian Nubert is a final-year bachelor student in electrical engineering at the ETH Zurich and has his responsibilities in the Localisation \& Mapping, State Estimation and Sensors \& Electronics in Project-ARC.
\end{minipage}
&
\begin{minipage}[b]{21mm}
\includegraphics[width=20mm]{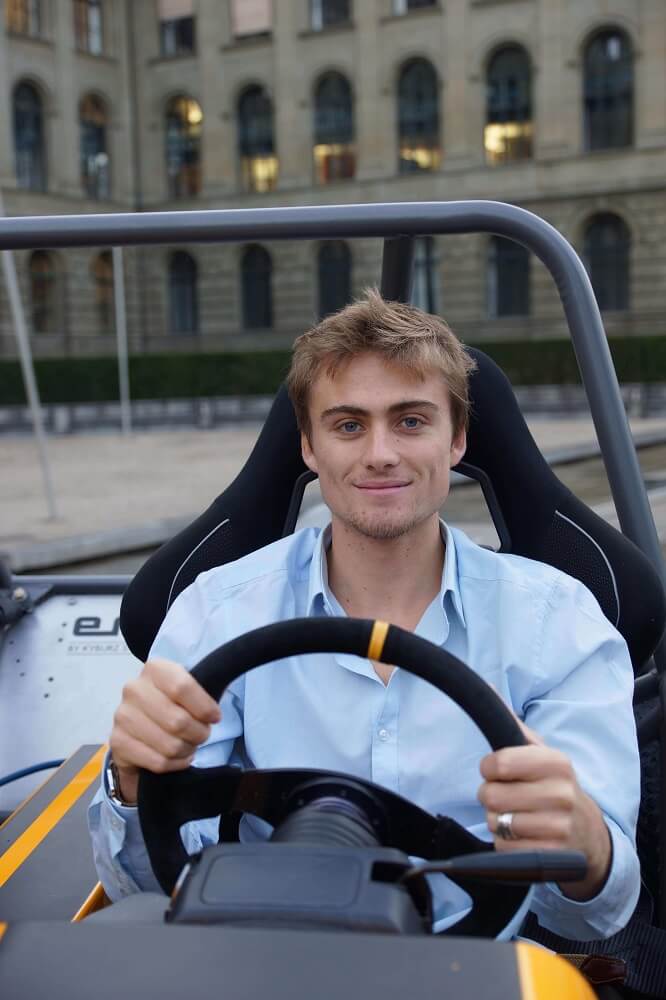}
\end{minipage} 
&
\begin{minipage}[b]{60mm}
\small
Moritz Patriarca is a mechanical engineering student in his final year at ETH Zurich . In the project ARC he is responsible for control, actuation and safety.
\end{minipage} 
\\
\end{tabular}
\vspace{1cm}

\begin{tabular}{cccc}
\begin{minipage}[b]{21mm}
\includegraphics[width=20mm]{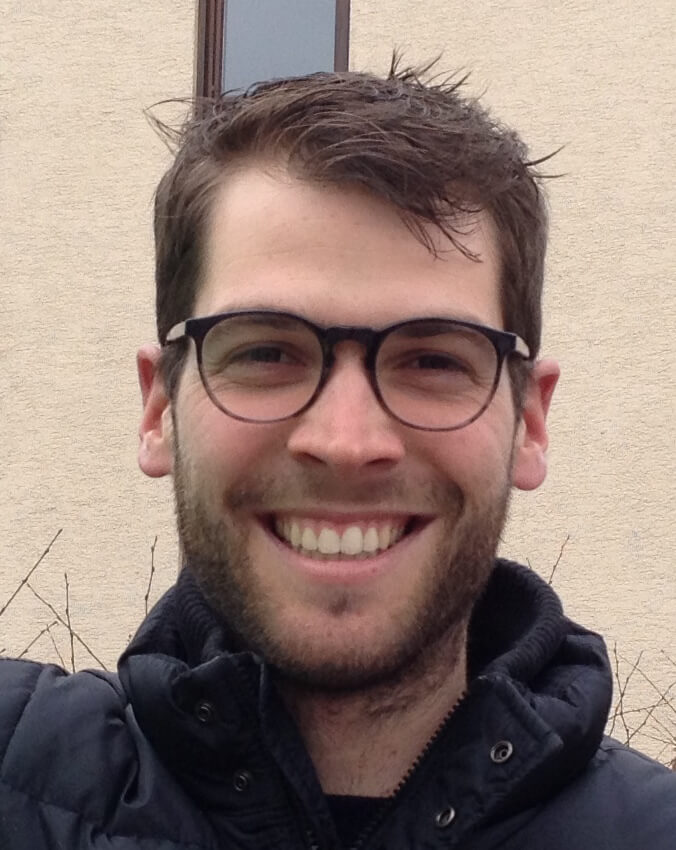}
\end{minipage} 
& 
\begin{minipage}[b]{60mm}
\small
Mark Pfeiffer completed both his Bachelor and Master degree at ETH Zurich. After having completed his Masters thesis at UC Berkeley he is currently pursuing his PhD at the Autonomous Systems Lab at ETH Zurich, focusing on motion planning and leveraging learned motion models for ground robot navigation.
\end{minipage}
&
\begin{minipage}[b]{21mm}
\includegraphics[width=20mm]{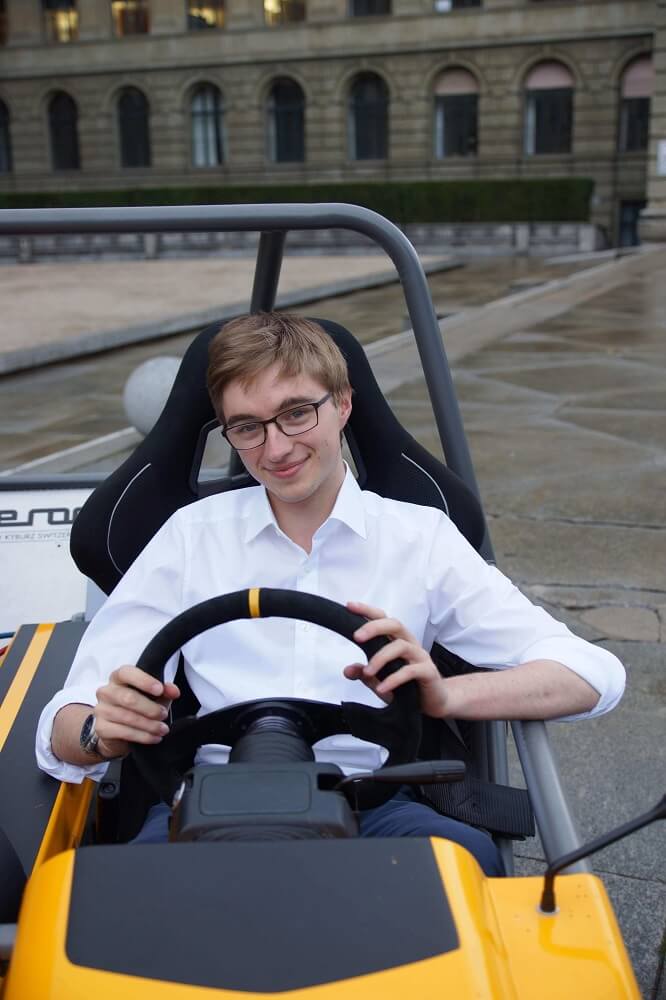}
\end{minipage} 
&
\begin{minipage}[b]{60mm}
\small
Simon Schaefer studies mechanical engineering BSc at the ETH Zurich, focussing Robotics and Control. In Project ARC he is responsible for State Estimation and High-Level interfaces as well as team leader. 
\end{minipage} 
\\
\end{tabular}
\vspace{1cm}

\begin{tabular}{cccc}
\begin{minipage}[b]{21mm}
\includegraphics[width=20mm]{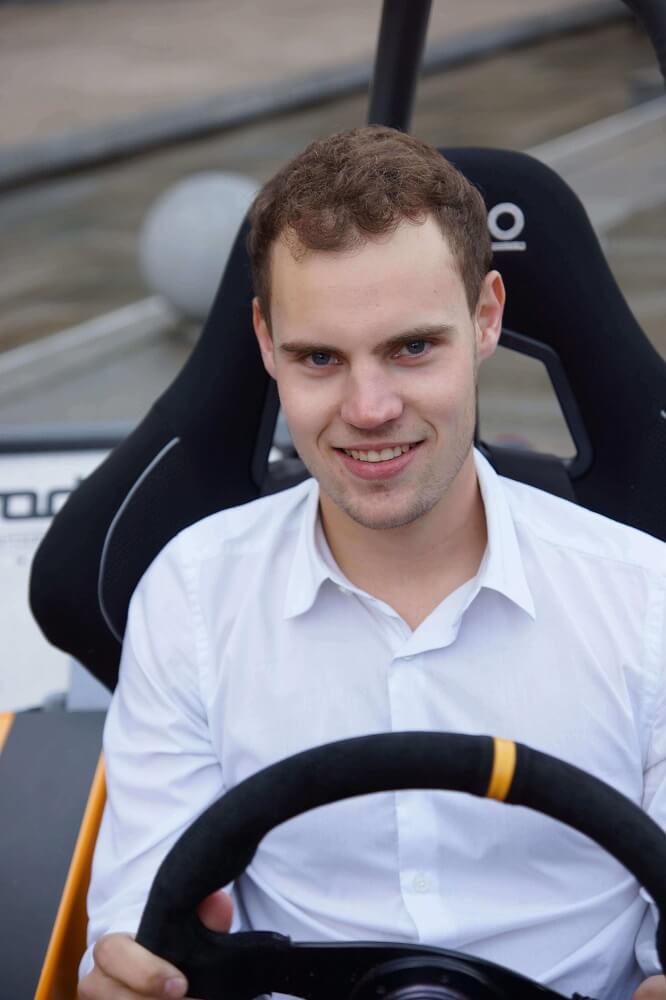}
\end{minipage} 
& 
\begin{minipage}[b]{60mm}
\small
Dominic Scotoni studies mechanical engineering BSc in the 6th semester at the ETH Zurich. He was responsible for the construction of the sensor mount and the electric steering \& braking system as well as the low-level velocity control of the car. In his bachelor thesis he developed a model based launch control to maximise the acceleration and stability of the car. 
\end{minipage}
&
\begin{minipage}[b]{21mm}
\includegraphics[width=20mm]{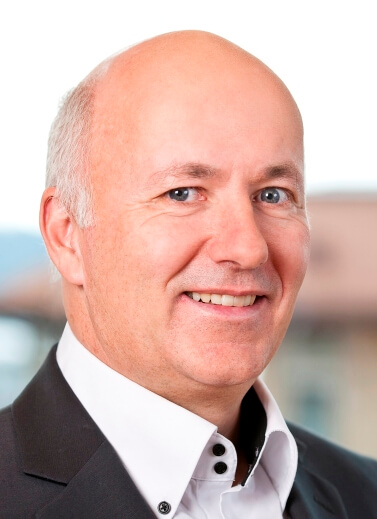}
\end{minipage} 
&
\begin{minipage}[b]{60mm}
\small
Roland Siegwart (born in 1959) is director of the Autonomous Systems Lab (ASL) of the Institute of Robotics and Intelligent Systems at ETH Zurich in Switzerland and a well known robotics expert. He was born in Lausanne and grew up in the Canton of Schwyz.
\end{minipage} 
\\
\end{tabular}
\vspace{1cm}

\begin{tabular}{cc}
\begin{minipage}[b]{21mm}
\includegraphics[width=20mm]{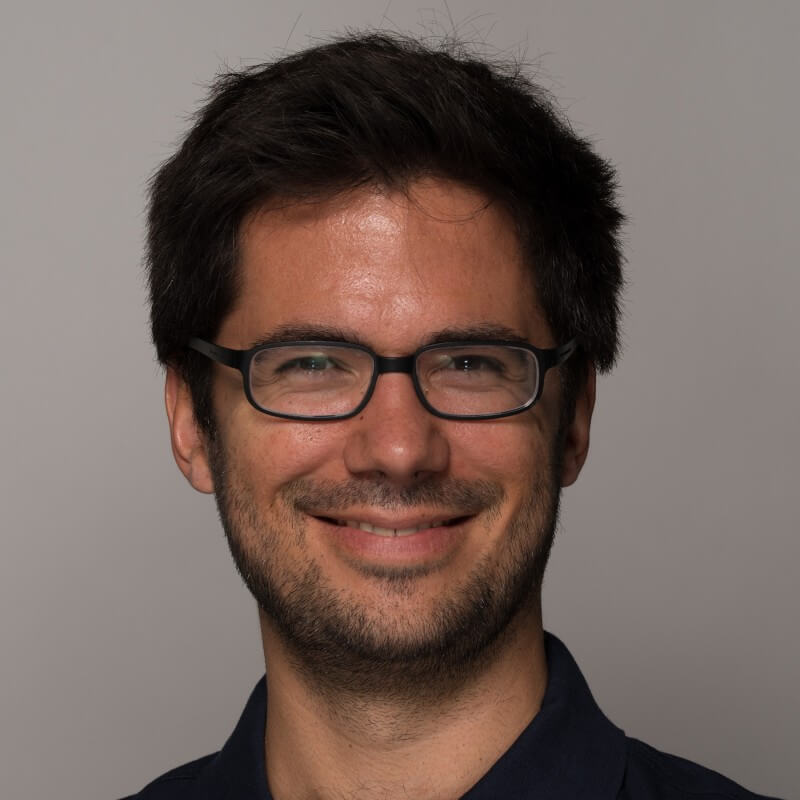}
\end{minipage} 
& 
\begin{minipage}[b]{60mm}
\small
Erik Wilhelm earned his Bachelor and Master in Chemical Engineering at the University of Waterloo. His Dr. Sci-ETH Zurich was conferred in Vehicle Design with an emphasis on mathematical modeling and simulation in 2011. He is currently the Lead Data Scientist at Kyburz Switzerland.
\end{minipage}
\\
\end{tabular}

\end{document}